%% file: arxiv.tex
\documentclass[prx,aps,showpacs,twocolumn,superscriptaddress,longbibliography,floatfix,10pt,nofootinbib]{revtex4-2}
\usepackage{graphicx}
\usepackage{mathrsfs}
\usepackage{amssymb}
\usepackage{amsmath}
\usepackage{physics}
\usepackage{bbding}
\usepackage{xcolor}
\usepackage{booktabs}
\usepackage{enumitem}
\usepackage{bm}
\usepackage[colorlinks=true,linkcolor=blue,anchorcolor=red,citecolor=blue,urlcolor=blue]{hyperref}
\DeclareMathAlphabet\mathbfcal{OMS}{cmsy}{b}{n}

\begin{document}

\title{FLARE: Flow Matching with Local Axis-Angle Representations for Stochastic Micromagnetic Evolution}

\author{Pengyu Li}
\email{These authors contributed equally to this work.}
\affiliation{State Key Laboratory of Quantum Functional Materials, Department of Physics, and Guangdong Basic Research Center of Excellence for Quantum Science, Southern University of Science and Technology, Shenzhen 518055, China}
\author{Renjie Tong}
\email{These authors contributed equally to this work.}
\affiliation{State Key Laboratory of Quantum Functional Materials, Department of Physics, and Guangdong Basic Research Center of Excellence for Quantum Science, Southern University of Science and Technology, Shenzhen 518055, China}
\author{Xuanlue Jiang}
\email{These authors contributed equally to this work.}
\affiliation{State Key Laboratory of Quantum Functional Materials, Department of Physics, and Guangdong Basic Research Center of Excellence for Quantum Science, Southern University of Science and Technology, Shenzhen 518055, China}
\author{Jianmin Li}
\email{These authors contributed equally to this work.}
\affiliation{State Key Laboratory of Quantum Functional Materials, Department of Physics, and Guangdong Basic Research Center of Excellence for Quantum Science, Southern University of Science and Technology, Shenzhen 518055, China}
\author{Yuanyuan Zhou}
\email{zhouyy3@sustech.edu.cn}
\affiliation{State Key Laboratory of Quantum Functional Materials, Department of Physics, and Guangdong Basic Research Center of Excellence for Quantum Science, Southern University of Science and Technology, Shenzhen 518055, China}
\newcommand{\fix}{\marginpar{FIX}}
\newcommand{\new}{\marginpar{NEW}}

\begin{abstract}
Long-horizon micromagnetic simulation remains expensive because conventional and learned solvers typically propagate Landau--Lifshitz--Gilbert (LLG) dynamics step by step. Existing learned approaches generally retain stepwise integration or model deterministic evolution, leaving full-field, direct-horizon stochastic prediction largely unexplored. We propose FLARE, a flow-matching framework that recasts stochastic finite-time magnetization prediction as conditional transport over anchor-relative local axis-angle rotations. This rotation-space formulation respects the intrinsic geometry of magnetization dynamics and preserves pointwise unit norm by construction. By explicitly conditioning on the physical prediction horizon, FLARE directly generates full-field stochastic endpoints across multiple target times without stepwise integration.  Against the strongest single-checkpoint external baseline on each metric, FLARE achieves 29.9\% lower angular energy distance ($15.30^\circ$), and a 37.3\% lower fair energy score (0.393). On a representative composed 5-ns two-segment protocol, FLARE achieves a $3{,}062\times$ best-batch speedup over the widely used GPU micromagnetic solver MuMax$^3$ on a single GPU.
\end{abstract}
\maketitle

\section{Introduction}
Magnetization dynamics can be viewed as a structured stochastic generative problem: given a current magnetization field, physical controls, and a target horizon, the goal is to predict the distribution of possible future states. Conventional micromagnetic solvers obtain these outcomes by integrating the Landau--Lifshitz--Gilbert (LLG) equation through many small time steps, making long-horizon and ensemble simulation expensive~\citep{donahue1999oommf,vansteenkiste2014design,fert2017magnetic,
reichhardt2022statics,bruckner2023magnum,moreels2026mumaxplus}.

Learned models can amortize this cost, but existing approaches largely focus on specialized magnetic tasks, sequential surrogate simulation, or deterministic prediction~\citep{iakovlev2018supervised,deviatov2019recurrent,wang2021learning,
feng2024classification,cai2024fast,chern2026jmmm,chern2026machine}. This leaves direct generation of full-field stochastic endpoints across physical horizons largely unexplored in micromagnetics.

We address this problem with \textbf{FLARE}
, a conditional generative framework built around an anchor-relative rotational representation of magnetization change. Because each local magnetizaiton is contrained to the unit sphere, its finite-time change is naturally expressed relative to its current orientation. Rather than transporting absolute Cartesian future states, FLARE represents each transition as a field of local axis-angle rotations relative to the current magnetization. Conditional flow matching learns the distribution of these rotation fields~\citep{lipman2022flow}, while Rodrigues decoding preserves pointwise unit norm by construction. Explicit physical-horizon conditioning enables a single model to generate stochastic endpoints across target times without stepwise LLG integration. Generated endpoints can further serve as anchors for subsequent control segments.
Our contributions are:
\textbf{Anchor-relative rotation-field transport.} We formulate stochastic finite-time magnetization transitions in local rotational coordinates relative to the anchor and instantiate this representation with axis-angle rotation fields. Matched controls isolate its advantage over Cartesian transport.
 \textbf{Direct stochastic endpoint generation across horizons.}\ We introduce horizon-conditioned stochastic endpoint generation as an alternative to sequential physical-time propagation, and demonstrate generalization to held-out horizons and self-generated anchors under changing controls.
 \textbf{Multi-criteria evaluation framework.} We develop a unified evaluation framework for stochastic dynamics that jointly measures sample-level field error, ensemble distribution fidelity, physical-observable distributions, and complete-output latency. %

\section{Background and Related Work} 
\label{gen_inst} 
\textbf{Micromagnetic dynamics and numerical acceleration.} Classical micromagnetics represents magnetic systems through a continuum magnetization field \citep{Brown1963Micromagnetics,Abert2019Micromagnetics}. The dynamics of the normalized magnetization are commonly governed by the Landau--Lifshitz--Gilbert (LLG) equation \citep{landau1935theory,gilbert2004phenomenological}. At finite temperature, thermal fluctuations can be incorporated through stochastic field terms, leading to stochastic Landau--Lifshitz or LLG dynamics \citep{mentink2010stochastic,roma2014stochastic,leliaert2017stochastic}. Established packages such as OOMMF, MuMax$^3$, Fidimag, magnum.np, and MuMax+ numerically integrate these equations on CPU or GPU~\citep{donahue1999oommf,vansteenkiste2014design,bisotti2020fidimag,bruckner2023magnum,moreels2026mumaxplus,abert2025neuralmag}. 
Nevertheless, high-resolution, long-horizon, and ensemble simulations remain costly because  resolving the dynamics requires propagating the full magnetization field through many successive physical time steps.
\textbf{Learned simulators and neural operators.} Learned simulators reduce repeated simulation cost by approximating dynamical mappings from data, including graph- and mesh-based autoregressive models \citep{sanchez2020learning,pfaff2021learning,wu2024estag} and neural operators for PDE surrogate modeling \citep{lu2021learning,li2021fourier,kovachki2023neural,wu2024transolver}, as well as continuous-time, latent, and structure-preserving formulations of learned dynamics \citep{yin2023dino,azencot2020koopman,bruna2024galerkin,matsubara2020energy,carnazza2024manybody}; approximation-theoretic analyses further motivate neural surrogates for parametric PDEs \citep{kutyniok2019pde}. In micromagnetics, learning has been applied to magnetostatic fields, inverse problems, energy minimization, reduced-order dynamics, and low-dimensional observables \citep{kovacs2022pinn,schaffer2023physics,schaffer2024constraintfree,kovacs2019learning,exl2020learning,exl2021lowrank,schaffer2022prediction,chen2022forecasting,dolui2026node}. Hybrid approaches such as NeuralMAG and MagneX replace expensive field computations with neural surrogates while retaining sequential time integration \citep{cai2024fast,nonaka2026magnex}; PDE-Refiner improves the stability of long-horizon neural PDE rollouts through iterative refinement, while likewise retaining sequential temporal propagation \citep{lippe2023pderefiner}. Existing learned approaches have largely focused on deterministic state prediction, reduced-order evolution, or accelerating components of sequential integration, rather than conditional distribution modeling of full-field finite-temperature dynamics.

\textbf{Generative models and flow matching for stochastic dynamics.} 
Generative surrogates extend learned simulation from point prediction to distributions over possible physical outcomes, with applications spanning fluid dynamics, PDE forecasting, molecular systems, and weather\citep{kohl2023turbulence,andrae2024continuous,cachay2023dyffusion,raonic2026gencfd,kossaifi2026weather,fotiadis2025afm,zhou2025generating,li2026generative,huang2024diffusionpde}. Recent examples include TurbDiff for turbulent flows, GenCast for ensemble weather prediction, MDGEN for molecular-dynamics trajectories, and MarS-FM for Markov-state molecular transitions \citep{lienen2024turbulence,price2025gencast,jing2024generative,kapusniak2025marsfm}. Flow Matching learns a continuous-time vector field that transports a simple source distribution to a target data distribution \citep{lipman2022flow,kerrigan2024ffm}, and has been applied to finite-horizon atomic transport composed into longer trajectories and to DFT Hamiltonian prediction \citep{nam2025flow,kim2025qhflow}. 

These studies establish generative modeling as a viable surrogate paradigm for stochastic physical dynamics. Micromagnetics introduces additional geometric structure: the state is a full spatial vector field at finite temperature, with each local magnetization constrained to lie on the unit sphere. 
Related geometric generative models accommodate manifold constraints through Riemannian score-based and diffusion processes \citep{debortoli2022riemannian,huang2022riemannian,lou2023scaling}, Riemannian flow matching \citep{chen2024flow,zaghen2026rvfm,cheng2025rcm}, and Lie-group and structured-domain formulations \citep{zhu2024tdm,bose2024se3,miller2024flowmm,bertolini2025lie,li2026gauge}; generation of manifold-supported data has also been analyzed theoretically \citep{loaiza2024manifold,zhang2026manifold}.
FLARE instead models finite-time magnetization change in anchor-relative rotational coordinates, while retaining Euclidean flow matching and enforcing the spherical constraint exactly through rotation-based decoding. Unlike finite-horizon transport formulated in global Euclidean coordinates \citep{nam2025flow}, FLARE defines these rotational coordinates locally with respect to the anchor magnetization at each spatial site.
\begin{figure*}[t]
    \centering
    \includegraphics[width=\linewidth]{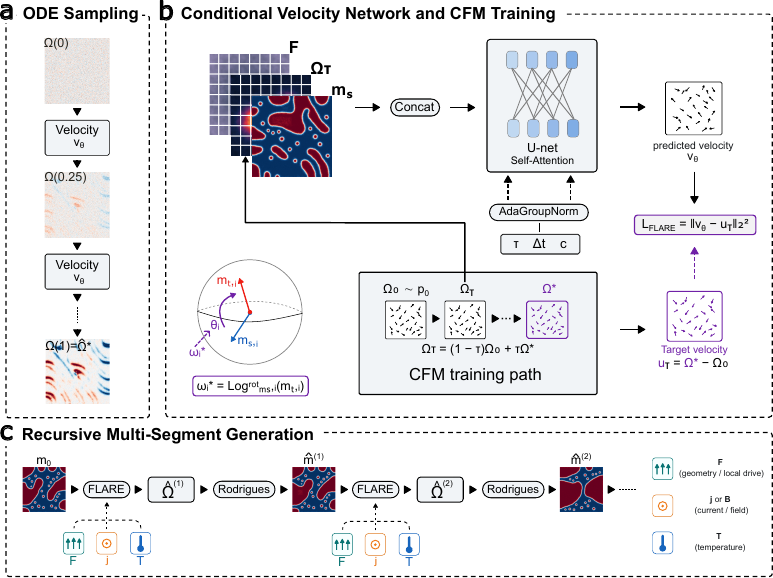}
    \caption{
Overview of FLARE.
\textbf{a}: During inference, ODE sampling transports an initial Gaussian rotation field $\Omega_0$ along the learned velocity field to the terminal rotation prediction $\widehat{\Omega}^{\star}=\Omega(1)$.
\textbf{b}: FLARE represents the transition from the anchor magnetization $\mathbf{m}_s$ to the target $\mathbf{m}_t$ as a local axis--angle rotation field $\Omega^\star$ and trains a conditional U-Net with self-attention using the CFM path $\Omega_\tau=(1-\tau)\Omega_0+\tau\Omega^\star$. Spatial conditions $F$ are concatenated with $\Omega_\tau$ and $\mathbf{m}_s$, while the flow time $\tau$, physical horizon $\Delta t$, and global conditions $c_g$ are injected through AdaGroupNorm.
\textbf{c}: The sampled terminal rotation field is mapped back to magnetization space through Rodrigues rotation. For multi-segment dynamics, each generated endpoint is recursively used as the anchor of the next FLARE call under the corresponding segment conditions.
}
    \label{fig:rotoFLARE_overview}
\end{figure*}

\section{FLARE}
\label{sec:magnet}

\subsection{Problem setup}
We consider a three-component unit magnetization field
$\mathbf{m}:\Omega \times [0,S] \to \mathbb{S}^2$
on a two-dimensional spatial domain $\Omega \subset \mathbb{R}^2$.
Its evolution is governed by the stochastic LLG equation~\citep{landau1935theory,gilbert2004phenomenological}
\begin{equation}
\begin{aligned}
\frac{\partial \mathbf{m}}{\partial t}
&= -\gamma\, \mathbf{m} \times H_{\mathrm{eff}}(\mathbf{m}; \eta, \xi_t)
\\
&\quad + \alpha\, \mathbf{m} \times \frac{\partial \mathbf{m}}{\partial t} + T(\mathbf{m}; u(t), \eta).
\end{aligned}
\label{eq:llg_reference}
\end{equation}
where $\|\mathbf{m}(\mathbf{r},t)\|_2=1$, $\mathbf{r}$ denotes spatial position and $t$ denotes the physical time. Here, $\gamma$ is the gyromagnetic ratio, $\alpha$ is the Gilbert damping coefficient, $\xi_t$ denotes finite-temperature stochastic forcing in $H_{\mathrm{eff}}$, $T$ is spin torque and $u(t)$ denotes controls such as external fields or currents, and $\eta$ collects static system parameters such as material properties and geometry.
Given an anchor magnetization field ($\mathbf{m}(t)$), a physical condition ($c$), and a target horizon ($\Delta\!t$), our goal is to model the conditional distribution:
\begin{equation}
    p\!\left(
\mathbf{m}{(t+\Delta\!t      )}
\mid
\mathbf{m}_t,c,\Delta\!t
\right),
\label{eq}
\end{equation}
where $c=(c_g,F)$ denotes the physical condition, with $c_g$ containing global or segment-level controls and $F$ encoding spatially resolved conditions such as geometry and local driving.
A single query therefore produces a \emph{future state} at the requested physical horizon. We use \emph{trajectory} exclusively for a time-resolved simulator evolution, \emph{segment endpoint} for a state at a control boundary, and \emph{generated state sequence} for recursively composed predictions.

Realistic simulation protocols may contain multiple stages with different physical controls. We represent such a protocol as $K$ conditionally defined segments with conditions $(c^{k})_{k=1}^K$, where each $c^{k}$ specifies the physical environment and duration of the $k$-th segment. Multi-stage generation is then defined recursively as:
\begin{equation}
\begin{gathered}
\hat{\mathbf{m}}^{(0)}=\mathbf{m}_0,\\
\hat{\mathbf{m}}^{(k)}
\sim
p_\theta\left(
\mathbf{m}
\mid
\hat{\mathbf{m}}^{(k-1)},c^{(k)}
\right),
\\
k=1,\ldots,K.
\label{eq:multisegment_generation}
\end{gathered}
\end{equation}
Each generated endpoint serves as the anchor for the subsequent segment, and the resulting sequence $(\hat{\mathbf{m}}^{(k)})_{k=1}^{K}$ forms a generated state sequence across the prescribed multi-stage protocol.
\subsection{Model formulation}

\label{sec:model_architecture}

\paragraph{Local Axis-Angle Transition Representation.}
The overall architecture and generation pipeline of FLARE are illustrated in Figure~\ref{fig:rotoFLARE_overview}. Each magnetic moment satisfies $\mathbf m_{s,i},\mathbf m_{t,i}\in S^2$. Rather than representing finite-time magnetization change as an unconstrained Cartesian displacement, we represent the transition from the anchor \(\mathbf m_{s,i}\) to the target \(\mathbf m_{t,i}\) by their principal axis-angle rotation
\citep{lewis2003geometric,daquino2005geometrical}:
\begin{equation}
\begin{gathered}
\theta_i
=
\operatorname{atan2}
\left(
\left\|
\mathbf m_{s,i}\times\mathbf m_{t,i}
\right\|_2,
\mathbf m_{s,i}^{\mathsf T}\mathbf m_{t,i}
\right),
\\
\boldsymbol{\omega}_i^\star
=
\theta_i
\frac{
\mathbf m_{s,i}\times\mathbf m_{t,i}
}{
\left\|
\mathbf m_{s,i}\times\mathbf m_{t,i}
\right\|_2
}.
\label{eq:rotation_vector}
\end{gathered}
\end{equation}
We denote this anchor-relative rotation encoding by
$\boldsymbol{\omega}_i^\star
=\operatorname{Log}^{\mathrm{rot}}_{\mathbf m_{s,i}}
(\mathbf m_{t,i})$.
For the ideal encoding,
\(\boldsymbol{\omega}_i^\star\perp\mathbf m_{s,i}\) and
\(\|\boldsymbol{\omega}_i^\star\|_2=\theta_i\in[0,\pi]\), so its direction specifies the local rotation axis and its magnitude the principal rotation angle.
Given $\theta=\|\boldsymbol{\omega}\|_2$ and
$\mathbf a=\boldsymbol{\omega}/\theta$, the corresponding rotation is
reconstructed through Rodrigues' formula
\citep{grassia1998practical,gallego2015compact}
\begin{equation}
\begin{gathered}
\mathcal R(\boldsymbol{\omega})
=
\mathbf I
+
\sin\theta[\mathbf a]_\times
+
(1-\cos\theta)[\mathbf a]_\times^2,
\\
\mathbf m_{t,i}
=
\mathcal R(\boldsymbol{\omega}_i^\star)\mathbf m_{s,i}.
\label{eq:rodrigues}
\end{gathered}
\end{equation}
Because \(\mathcal R(\boldsymbol{\omega})\in SO(3)\), the reconstruction preserves the pointwise unit norm of the magnetization by construction. For nearly identical moments, we use the first-order limit
$\boldsymbol{\omega}_i^\star
\approx\mathbf m_{s,i}\times\mathbf m_{t,i}$.
Near antipodal configurations, we adopt a deterministic tangent-axis $\pi$-rotation approximation. Implementation details and statistics for these corner cases are provided in Appendix~\ref{app:rotation}. The resulting rotation field $\Omega^\star=\{\boldsymbol{\omega}_i^\star\}_{i=1}^{N}$ serves as the representation on which the conditional flow model is defined.

\paragraph{Conditional Flow Matching in Rotation Space}
Given the endpoint rotation field
$\boldsymbol{\Omega}^\star$, we sample a Gaussian source \(\boldsymbol{\Omega}_0\sim p_0\) and a flow time $\tau\sim\mathcal U(0,1)$ and construct the straight
conditional path \citep{lipman2022flow,tong2024improving}
\begin{equation}
\boldsymbol{\Omega}_\tau
=
(1-\tau)\boldsymbol{\Omega}_0
+
\tau\boldsymbol{\Omega}^\star,
\qquad
\mathbf u_\tau
=\frac{\partial \boldsymbol{\Omega}_{\tau}}{\partial \tau}
=\boldsymbol{\Omega}^\star-\boldsymbol{\Omega}_0.
\end{equation}
Here, \(\tau\in[0,1]\) denotes the generative flow time and is distinct from the physical prediction horizon \(\Delta t\).
We learn a conditional vector field
$\mathbf v_\theta
\left(
\boldsymbol{\Omega}_\tau,\tau
\mid
\mathbf m_s,\Delta t,c_g,\mathbf F
\right)$
to approximate the target velocity \(\mathbf u_\tau\). The model is trained with a conditional flow-matching objective
\begin{equation}
\mathcal L_{\mathrm{FLARE}} =
\mathbb E
\left[
\left|
\mathbf v_\theta
\left(
\boldsymbol{\Omega}_\tau,\tau
\mid
\mathbf m_s,\Delta t,c_g,\mathbf F
\right)-
\mathbf u_\tau
\right|_2^2
\right],
\label{eq:rotation_cfm_loss}
\end{equation}
where the expectation is taken over training transitions, Gaussian source samples, and flow times. Uniform weighting is used across sampled flow times, prediction horizons, and driving categories.

In inference, we integrate the learned vector field in rotation space,
\begin{equation}
\frac{d\boldsymbol{\Omega}}{d\tau}=
\mathbf v_\theta
\left(
\boldsymbol{\Omega},\tau
\mid
\mathbf m_s,\Delta t,c_g,\mathbf F
\right),
\qquad
\boldsymbol{\Omega}(0)=\boldsymbol{\Omega}_0,
\label{eq:rotation_flow_ode}
\end{equation}
to obtain the terminal rotation field \(\widehat{\boldsymbol{\Omega}}^\star=\boldsymbol{\Omega}(1)\). The sampled rotation need not be the minimal one, since \(\|\widehat{\boldsymbol{\omega}}_i^\star\|_2\) may exceed \(\pi\), but Rodrigues decoding still returns a unit vector at every cell; see Eq.~\ref{eq:rodrigues}.

\paragraph{Conditional Vector-Field Network}

We parameterize \(\mathbf v_\theta\) with a U-Net at native spatial resolution. Its spatial input concatenates the anchor magnetization \(\mathbf{m}_s\), rotation state \(\Omega_\tau\), and full-resolution condition features \(F\), yielding 19 input channels. The feature maps \(\mathbf F\) encode spatially resolved information such as geometry and position-dependent driving fields. 

Global conditioning is handled separately. The flow time \(\tau\), target horizon \(\Delta t\), and segment-level physical controls \(c_g\) are embedded into a 512-dimensional conditioning vector, which modulates the U-Net features at each resolution through AdaGroupNorm. An eight-head self-attention block at the bottleneck captures long-range spatial dependencies. The network outputs a three-channel velocity field 
$
\mathbf v_\theta
\left(
\boldsymbol{\Omega}_\tau,\tau
\mid
\mathbf m_s,\Delta t,c_g,\mathbf F
\right),
$
which parametrizes the generative ODE in Eq.~\ref{eq:rotation_flow_ode}.

\paragraph{Training-Pair Construction}

We partition each trajectory at changes in the driving protocol and sample training pairs within individual constant-control segments. After uniformly sampling a valid trajectory and target type, we sample a valid anchor \(s\) and candidate horizon \(h\); the target is the saved frame nearest \(s+h\) that remains within the same segment. The actual timestamp difference defines the physical horizon, \(\Delta t=t_{\mathrm{target}}-t_{\mathrm{anchor}}\), yielding a training tuple \((\mathbf m_s,\mathbf m_t,\Delta t,c_g, F)\), where \(c_g\) denotes the corresponding segment conditions. For validation, we use the earliest valid anchor in each admissible range, yielding a deterministic protocol aligned with rollout from the segment start.

\begin{figure*}[t]
    \centering
    \includegraphics[width=\linewidth]{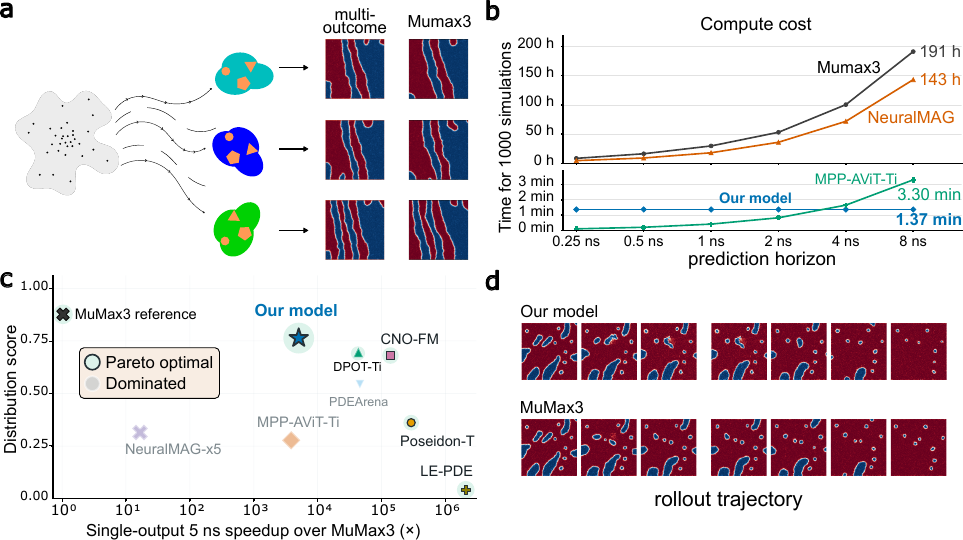}
    \caption{Overview of the task, generated dynamics, and empirical performance.  \textbf{a}: The multi-outcome effects of our model.  \textbf{b}: runtime-only comparison on one H200. \textbf{c}: Pareto-optimal plot of the speeds and distribution scores for different models. \textbf{d}: the corresponding results from the MuMax$^3$ simulation, showing representative multi-segment magnetization outcomes.}
    \label{fig:main_overview}
\end{figure*} 

\subsection{Distributional Evaluation Metric.}\label{sec:distribution_metric}
Magnetization dynamics are stochastic, so agreement with a single MuMax$^3$ realization is insufficient to assess whether a generative model captures the distribution of physically plausible outcomes. We therefore define a distributional metric that compares sets of magnetization fields while accounting for small spatial displacements of otherwise similar magnetic textures. 

\begin{figure*}[t]
\centering
\includegraphics[width=0.5\linewidth]{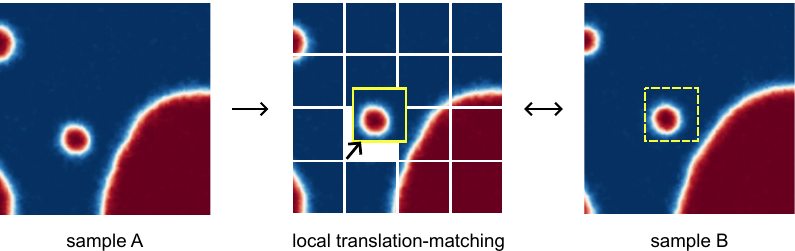}
\caption{Illustration of the local translation-matching procedure. The $256\times256$ magnetization field is divided into $4\times4$ local patches, and each patch is independently matched within a translation range of $\pm16$ grid cells along each spatial direction.}
\label{fig:local_translation_matching}
\end{figure*}

\paragraph{Local translation-matching distance} Direct cell-wise distance can strongly penalize small translations of domain walls, 
vortices, or other localized structures. To obtain a more structure-aware comparison, we adopt a local translation-matching procedure inspired by displacement search in image registration and block matching  \citep{jain1981displacement,zitova2003registration,barnes2009patchmatch}. As illustrated in Figure~\ref{fig:local_translation_matching}, the magnetization field is partitioned into local patches, and each patch is independently matched over a bounded translation window. 
For normalized magnetization fields $\mathbf m_a$ and $\mathbf m_b$, we define
\begin{equation}
\begin{gathered}
C_p^{a\to b}=\max_{\Delta r\in\mathcal W}\frac{1}{|\Omega_p|}
\sum_{r\in\Omega_p}m_a(r)\!\cdot\!m_b(r+\Delta r),
\\
D(a,b)=\frac{1}{2P_{\rm patch}}\sum_{p=1}^{P_{\rm patch}}
\left(2-C_p^{a\to b}-C_p^{b\to a}\right).
\label{eq:local_translation_distance}
\end{gathered}
\end{equation}
where \(\Omega_p\) denotes the spatial support of patch \(p\) and \(\mathcal W\) is the allowed translation window. Locations shifted outside the magnetic geometry contribute zero. The symmetric distance \(D(a,b)\) therefore measures disagreement in the local magnetic structure while remaining tolerant to small spatial translations.

\paragraph{Distribution Score.} For a fixed physical condition, let
$X=\{x_i\}_{i=1}^{M}$ and $Y=\{y_j\}_{j=1}^{N}$ denote the reference MuMax$^3$ outcomes and the model outcomes, respectively.
We convert pairwise distance $D$ into a Gaussian similarity. For each MuMax$^3$ outcome $x_i$, we then estimate two local densities: one induced by the model-generated samples and a leave-one-out reference density estimated from the remaining MuMax$^3$ repeats. The kernel bandwidth is determined solely from nearest-neighbor distances among the MuMax$^3$ samples, so that the metric does not depend on the model outputs. The complete kernel and density definitions are provided in Appendix~\ref{app:metrics}.

Using these densities, the per-condition agreement score and the final Distribution Score are
\begin{equation}
\begin{aligned}
S_c
&=
\exp\left[
-\frac{1}{M}
\sum_{i=1}^{M}
\left|
\log
\frac{
q_{\mathrm{model}}(x_i)+\epsilon
}{
q_{\mathrm{MuMax}}^{-i}(x_i)+\epsilon
}
\right|
\right],
\\
S_{\mathrm{dist}}
&=
\exp\left[
\frac{1}{N_{\mathrm{cond}}}
\sum_{c=1}^{N_{\mathrm{cond}}}
\log S_c
\right],
\end{aligned}
\label{eq:distribution_score}
\end{equation}
where $\epsilon=10^{-12}$. The resulting distribution score lies in \((0,1]\), with values closer to one indicating better agreement between the generated and MuMax$^3$ endpoint distributions.

\section{Experiments}
\label{sec:experiments}

\subsection{Experimental Setup}
\label{sec:exp_setup}

\paragraph{Dataset and splits.}
We evaluate FLARE on a finite-temperature micromagnetic dataset generated with MuMax$^3$, comprising 7,770 completed driven and post-drive-relaxation trajectories from 1,000 base conditions. Each state is a three-component magnetization field on a $256\times256\times1$ finite-difference grid corresponding to a $512\times512\times1$~$\mathrm{nm}^3$ magnetic film with open boundaries. The configured 1--6~ns interval refers to absolute simulation times rather than one-shot prediction horizons.  The material parameters are fixed across the corpus and include interfacial Dzyaloshinskii--Moriya interaction, while temperature, out-of-plane field, current amplitude, initial texture, and the spatial current-mask family are varied. Detailed material parameters, control ranges, temperature levels, and the distribution over the five mask families are given in Appendix~\ref{app:implementation_details}. 
To avoid leakage between trajectories generated under the same physical setting, we partition by base condition, resulting in 6,227 training, 786 validation, and 757 test trajectories.

\paragraph{Comparators and execution protocol.}
We compare FLARE with author-released CNO~\citep{raonic2023cno}, Poseidon~\citep{herde2024poseidon}, DPOT~\citep{hao2024dpot}, MPP~\citep{mccabe2024multiple}, PDEArena~\citep{gupta2022towards}, LE-PDE~\citep{wu2022learning}, and NeuralMAG~\citep{cai2024fast}. These external comparators are deterministic per checkpoint, except for NeuralMAG-x5, which retains the thermal field; the two matched internal controls change only the transport coordinates or the learning objective. All learned models are adapted to the same x5 training split for 50,000 optimization updates and use the final checkpoint. Quality follows the exact-control complete path: each method predicts the driven segment from the saved pre-drive state, passes its own boundary prediction into the exact 3.5-ns relaxation, and is scored only at the final endpoint. No MuMax$^3$ state is injected at the handoff. Because complete-path durations vary, timing uses a separate standardized fixed two-segment 5-ns workload: a 2-ns driven segment followed by a 3-ns zero-current relaxation, with the first segment's model prediction handed to the second.
Each method keeps its native temporal interface. We report both batch-one latency and the minimum per-output latency over a power-of-two batch sweep; MuMax$^3$ is timed in its standard single-process execution mode, where $B{=}1$ is its fastest per-output setting.

\paragraph{Metrics and uncertainty.}
Distribution fidelity is evaluated over 33 held-out complete-path conditions, each with five repeat trajectories.
We use complementary diagnostics comprising the Distribution Score defined in Sec. \ref{sec:distribution_metric}, angular energy distance, and the proper fair energy score \citep{gneiting2007scoring,ferro2014fair}. Distribution Score and angular energy distance compare the five final model endpoints with the five corresponding MuMax\(^{3}\) endpoints as outcome sets, while paired angle matches repeat \(i\) and is retained as a realization-level diagnostic. Paired angle is minimized by a deterministic central estimate of the conditional distribution and equals the MuMax\(^{3}\) repeat-to-repeat level for a correct sample.
The proper fair energy score uses 165 exact drive-start anchors (33 bases $\times$ five repeats), with five forecasts per anchor for stochastic methods and a one-point forecast distribution for deterministic single checkpoints.
Because the primary x5 set contains only one MuMax$^3$ continuation per exact drive-start anchor, independent final-endpoint reference levels are estimated from the separate x30 audit (Appendix Table~\ref{tab:app_mumax_reference}), which provides an absolute scale. We report 95\% base-condition-clustered bootstrap intervals, retaining all five complete paths within each resampled base group. Angular error is the mean full-vector angular deviation, in degrees, over valid magnetic cells. The physical forecast spans range from 3.75 to 4.42 ns (mean 4.10 ns), reflecting the realized driven-segment durations followed by the exact 3.5-ns relaxation. Further resampling details are given in Appendix~\ref{app:implementation_details}.

\subsection{Main Results}
\label{sec:main_results}

\begin{table*}[t]
\centering
\scriptsize
\setlength{\tabcolsep}{2.2pt}
\caption{Single-checkpoint exact-control complete-path future-state quality; standardized fixed two-segment 5-ns timing. Every learned row uses 50,000 x5 updates; NeuralMAG-x5 retains the thermal field during both training and inference. Brackets are 95\% base-group cluster-bootstrap intervals. External rows are single checkpoints; Table~\ref{tab:app_seed_geometry} reports five-seed ensembles of the same baselines, where FLARE's fair-energy margin is 21.8\%. The top row is a simulator-only reference (17-base $\times$30 audit; Appendix Table~\ref{tab:app_mumax_reference}): a descriptive scale, not a ranked entry.}
\label{tab:main_distribution}
\resizebox{\linewidth}{!}{%
\begin{tabular}{lrrrrrrr}
\toprule
Method & $B{=}1$ latency (ms) & Distribution Score $\uparrow$ & Paired angle ($^\circ$) $\downarrow$ & Angular energy dist. ($^\circ$) $\downarrow$ & Fair energy score $\downarrow$ & $B{=}1$ speedup $\uparrow$ & Best-batch speedup $\uparrow$ \\
\midrule
\multicolumn{8}{l}{\emph{Reference}} \\
MuMax$^3$ 5 vs 5 & -- & 0.862 [0.801, 0.915] & 30.25 [23.89, 36.66] & 12.20 [9.62, 14.86] & -- & -- & -- \\
\midrule
\textbf{FLARE ODE-10} & 995.2 & \textbf{0.762} [0.660, 0.844] & 36.72 [31.93, 41.38] & \textbf{15.30 [13.32, 17.26]} & \textbf{0.393} [0.356, 0.430] & $448.3\times$ & $3{,}062.4\times$ \\
Poseidon-T & 61.91 & 0.341 [0.077, 0.749] & 35.60 [30.60, 41.11] & 25.05 [17.86, 34.56] & 0.751 [0.672, 0.830] & $7{,}206.3\times$ & $686{,}423.6\times$ \\
CNO-FM & 72.89 & 0.695 [0.577, 0.787] & \textbf{34.43 [30.41, 38.33]} & 21.82 [17.94, 26.23] & 0.745 [0.682, 0.807] & $6{,}120.7\times$ & $133{,}012.0\times$ \\
DPOT-Ti & 91.13 & 0.707 [0.599, 0.803] & 39.81 [34.46, 45.59] & 27.02 [19.98, 36.58] & 0.814 [0.735, 0.893] & $4{,}895.3\times$ & $48{,}327.7\times$ \\
MPP-AViT-Ti & 613.4 & 0.290 [0.059, 0.764] & 49.41 [43.24, 56.21] & 42.10 [32.57, 52.93] & 0.930 [0.849, 1.015] & $727.3\times$ & $3{,}730.7\times$ \\
PDEArena U-Net & 40.17 & 0.597 [0.394, 0.779] & 37.26 [32.48, 42.16] & 24.03 [19.43, 28.99] & 0.774 [0.697, 0.850] & $11{,}106.1\times$ & $45{,}720.7\times$ \\
LE-PDE & 7.642 & 0.0018 [0.0001, 0.0293] & 76.18 [70.31, 81.78] & 77.96 [65.49, 91.87] & 1.269 [1.207, 1.328] & $58{,}379.6\times$ & $\mathbf{1{,}749{,}670.6\times}$ \\
NeuralMAG-x5 & 335,786 & 0.316 [0.068, 0.731] & 53.54 [47.41, 59.54] & 36.76 [29.05, 45.58] & 0.627 [0.562, 0.690] & $1.33\times$ & $14.2\times$ \\
\midrule
\multicolumn{8}{l}{\emph{Matched internal controls}} \\
FLARE (Cartesian) & -- & 0.711 [0.563, 0.821] & 39.88 [35.54, 44.21] & 17.91 [15.52, 20.48] & 0.420 [0.383, 0.456] & -- & -- \\
Direct U-Net & -- & 0.477 [0.190, 0.783] & 38.53 [34.44, 42.76] & 21.41 [18.17, 24.91] & 0.808 [0.745, 0.867] & -- & -- \\
\bottomrule
\end{tabular}
}%
\vspace{2pt}
\parbox{0.98\linewidth}{\footnotesize Quality follows the protocol above. Cartesian CFM and Direct U-Net are internal controls: one changes coordinates; the other directly regresses $\Omega_\star$. All three use seed 78. Because quality-path durations vary, timing uses a fixed two-segment representative 5-ns workload (2 ns drive plus 3 ns relaxation, with one predicted-state handoff): $B=1$ is single-query latency, best batch is the fastest supported power of two, and speedups use the 446.125-s MuMax$^3$ reference. Neither control has timing; Table~\ref{tab:design_ablation}(a) gives the three-seed summary; Appendix Table~\ref{tab:app_seed_geometry} reports five-seed ensembles.}
\end{table*}

In Table~\ref{tab:main_distribution}, FLARE leads all seven author-released baselines on the three distributional metrics. Its Distribution Score is 0.762 versus 0.707 for DPOT-Ti (+7.9\%); its angular energy distance is 15.30$^\circ$ versus 21.82$^\circ$ for CNO-FM (29.9\% lower); and its fair energy score is 0.393 versus 0.627 for NeuralMAG-x5 (37.3\% lower). CNO-FM obtains a lower repeat-paired angle, 34.43$^\circ$ versus 36.72$^\circ$; this metric scores one indexed stochastic realization rather than the unordered outcome distribution. Matched Cartesian CFM and Direct U-Net test the rotation coordinates and flow-matching objective, respectively.
FLARE also leads five-model ensembles in distributional fidelity (Table~\ref{tab:app_seed_geometry}). FLARE achieves a $3{,}062\times$ best-batch speedup (448× at B=1) over MuMax$^3$ on the 5-ns workload
(Table~\ref{tab:main_distribution}).
On the driven segment (the only one with an exact-anchor oracle), FLARE's fair energy score is 0.322 [0.290, 0.353] versus 0.283 [0.230, 0.335] for the MuMax$^3$ reference. Single-segment diagnostics are reported in Appendices~\ref{app:distribution_observables}--\ref{app:metric_pressure}.
\textbf{Multi-segment Composition.} FLARE composes conditional endpoint distribution across changing controls by using each generated endpoint as the anchor for next segment, without intermediate LLG states. We evaluate both ground-truth-boundary and autoregressive handoffs over 736 rollouts from 97 held-out conditions. Predicted-state exposure improves autoregressive robustness over Stage 1 but does not outperform an equal-budget ground-truth-only continuation; we therefore treat it as an ablation rather than a separation contribution. Full results are provided in Appendix~\ref{app:composition} and Table~\ref{tab:app_stage2}.
\subsection{Out-of-Distribution Generalization}
\label{sec:ood}

We evaluate OOD generalization beyond the training distribution along two complementary axes: unseen spatial
current-mask family and unseen temperature.

\paragraph{Unseen current-mask family.}
We train a dedicated Stage~1 FLARE model after excluding all ring-current-mask
conditions from training and validation. On the held-out ring-mask family,
single-segment and autoregressive angular errors increase by
$0.91^\circ$ (3.12\%) and $0.95^\circ$ (3.04\%), respectively, relative to the in-distribution reference. This indicates
stable transfer to an unseen current-mask family. Full results are reported in
Appendix~\ref{app:ood}.

\paragraph{Unseen temperature.}
Using the frozen Stage~1 checkpoint trained only at 30, 150, and 300~K, we
evaluate new zero-shot test sets at 90 and 225~K. Relative to paired
interpolation references from the adjacent training temperatures,
single-segment and autoregressive angular errors change by $+1.27\%$ and
$-4.15\%$ at 90~K, and by $+1.64\%$ and $+2.37\%$ at 225~K. No evaluated
metric shows statistically resolved adverse degradation at either temperature,
showing robust interpolation to unseen in-range physical conditions. Full
metrics, confidence intervals, and the 400~K extrapolation stress test are
reported in Appendix~\ref{app:temperature_ood}.

\subsection{Ablation Studies}
\label{sec:ablation}

We isolate four design choices with matched controls: learned endpoint dynamics, anchor-relative rotational coordinates, physical-horizon conditioning, and the flow-matching objective (Table~\ref{tab:design_ablation}; Appendix~\ref{app:seed_geometry}). Persistence is a no-dynamics sanity check; learned controls share the architecture and training budget unless stated otherwise. The compute-matched rollout-state control is reported separately in Appendix~\ref{app:composition}.

\begin{table*}[t]
\centering
\small
\setlength{\tabcolsep}{3pt}
\caption{Stage~1 ablations. (a,b) Exact-control complete paths; (c) single-segment time interpolation. Panel (a) reports mean $\pm$ sample SD over seeds 78--80; (b,c) report seed-78 point estimates.}
\label{tab:design_ablation}
\resizebox{\linewidth}{!}{%
\begin{tabular}{lcccc}
\toprule
Variant & Distribution Score $\uparrow$ & Angular energy dist. ($^\circ$) $\downarrow$ & Paired angle ($^\circ$) $\downarrow$ & Fair energy score $\downarrow$ \\
\midrule
\multicolumn{5}{l}{\textit{(a) Representation and objective controls: three training seeds}} \\
FLARE & $\boldsymbol{0.7592\pm0.0056}$ & $\boldsymbol{15.31\pm0.24}$ & $\boldsymbol{36.70\pm0.23}$ & $\boldsymbol{0.3944\pm0.0028}$ \\
FLARE(Fixed 2D tangent) & $0.7562\pm0.0221$ & $16.70\pm2.09$ & $37.94\pm1.41$ & $0.3991\pm0.0155$ \\
Cartesian CFM & $0.7176\pm0.0073$ & $17.98\pm0.12$ & $39.96\pm0.29$ & $0.4210\pm0.0033$ \\
Riemannian FM & $0.7341\pm0.0071$ & $18.00\pm0.19$ & $39.01\pm0.10$ & $0.4330\pm0.0029$ \\
Direct U-Net & $0.4741\pm0.0040$ & $21.26\pm0.13$ & $38.60\pm0.07$ & $0.8087\pm0.0013$ \\
\bottomrule
\end{tabular}}
\par\vspace{3pt}
{\small
\begin{tabular*}{\linewidth}{@{\extracolsep{\fill}}llccc@{}}
\toprule
\multicolumn{5}{l}{\textit{(b) Component ablations}} \\
Design & Metrics & FLARE & Control & Gain \\
\midrule
Flow transport & Angle ($^\circ$) / MSE & \textbf{36.74} / \textbf{0.6127} & 67.03 / 1.4045 & +30.29 / +0.7918 \\
Rotation coords. & Angle ($^\circ$) / $|\Delta Q|$ & \textbf{36.74} / \textbf{0.7715} & 39.88 / 0.8786 & +3.15 / +0.1071 \\
Target time & Angle ($^\circ$) / $|\Delta\mathcal{E}|$ & \textbf{36.74} / \textbf{0.00977} & 47.16 / 0.01145 & +10.42 / +0.00167 \\
\midrule
\multicolumn{5}{l}{\textit{(c) Time interpolation}} \\
1.5 / 2.5 ns & Angle ($^\circ$) & \textbf{30.88} / \textbf{32.67} & 39.79 / 40.97 & +8.91 / +8.31 \\
\bottomrule
\end{tabular*}}
\vspace{2pt}
\parbox{0.98\linewidth}{\footnotesize (a,b): 33 bases $\times$ five endpoints, predicted drive-to-relaxation handoff, 3.5-ns relaxation; fair energy uses 165 anchors with five stochastic draws or one deterministic forecast. (b) Controls: persistence, Cartesian CFM, no-time. (c): 256 cases per horizon, true segment starts, no-time control.}
\end{table*}

As shown in Table~\ref{tab:design_ablation}(b), the persistence comparison confirms that FLARE learns conditional dynamics beyond copying the anchor. Both the 3D axis--angle and fixed 2D tangent outperform absolute-spin Cartesian CFM and Riemannian FM across all four metrics (Table~\ref{tab:design_ablation}(a)), associating the gain with rotational coordinates rather than dimensionality. We use the 3D form to avoid per-spin tangent frames and reduce seed variability. Target-horizon conditioning improves both complete-path predictions and interpolation at the held-out 1.5- and 2.5-ns horizons.

\section{Discussion and Limitations}

FLARE provides evidence that anchor-relative rotational coordinates are effective for modeling finite-temperature micromagnetic dynamics as conditional future-state distributions. At a matched five-sample budget, FLARE closely reproduces several physical-observable distributions, including topological charge and mean out-of-plane magnetization. FLARE also supports prediction-only composition across changing control segments, recursively using generated endpoints as subsequent anchors. Three-seed complete-path results favor the core model over Cartesian CFM, Riemannian FM, and Direct U-Net, while the ablations confirm the importance of explicit target-horizon conditioning. FLARE's advantage is distributional fidelity for stochastic future-state prediction: CNO-FM achieves a lower paired angular error, and several deterministic surrogates are substantially faster at inference.

Several limitations remain. First, sequential composition introduces additional error under predicted-state handoffs, and mixed rollout-state exposure does not outperform the equal-budget GT-only continuation, leaving extended multi-segment composition sensitive to accumulated handoff error.
Second, the present dataset covers a fixed material system and a limited range of temperatures, fields, currents, initial textures, and control geometries, so our results speak to that regime; extending it across materials and dynamical regimes requires new training data. Third, our experiments are restricted to two-dimensional thin-film micromagnetics and finite prediction horizons. Extension to three-dimensional systems, broader material and control distributions, and larger-scale training data is a natural direction for future work.

\bibliography{ref}

\clearpage
\appendix
\input{appendix}
\end{document}

%% file: appendix.tex
\makeatletter
\setlength{\@fptop}{0pt}
\makeatother

\section{Implementation and evaluation protocol}
\label{app:implementation_details}

This appendix is organized around the questions needed to reproduce and
stress-test the main claims.  We first specify the representation, model,
training procedure, split unit, and statistical unit.  We then report
distribution, solver, temporal, composition, and geometry-shift diagnostics.
The final part is a visual audit of the magnetization fields.  Every
qualitative case is selected by a numerical quantile fixed before rendering;
none is chosen by visual inspection.

\subsection{Rotation-space transition parameterization}
\label{app:rotation}

For each valid magnetic cell, let $a=m_s$ and $b=m_t$ be the unit magnetization at the anchor and target.  Ideally, FLARE represents their change by the minimum local rotation vector and decodes a predicted rotation with Rodrigues'
map:
\begin{equation}
\begin{aligned}
\omega^\star
&=\operatorname{atan2}(\|a\times b\|,a^\top b)
  \frac{a\times b}{\|a\times b\|},\\
\mathcal R(\omega)a
&=a\cos\theta+(u\times a)\sin\theta
  \\ &\quad +u(u^\top a)(1-\cos\theta),
\end{aligned}
\label{eq:app_rotation}
\end{equation}
where $\theta=\|\omega\|$ and $u=\omega/\theta$.  The implementation uses
the continuous small-angle limit. For $a^\top b<-0.999$, it uses a $\pi$ rotation about the tangent-plane projection of $e_z$ (or $e_x$ when $|a_z|>0.9$). This decodes to $-a$, approximating near-antipodal targets with angular error below $2.563^\circ$ while preserving unit norm.
Dot products are clipped before evaluating the angle.  Geometry masking is applied before constructing the target and after decoding.

Rodrigues' map is orthogonal, so a unit anchor remains unit length for any
finite predicted rotation.  A final normalization removes mixed-precision
roundoff; it is not used to repair an unconstrained Cartesian prediction.
This distinction explains the near-machine-precision norm error of the
rotation models.

The implementation uses the deterministic tangent-axis branch when
$a^\top b<-0.999$, equivalently $\theta>177.437^\circ$. Across the 66 primary
conditions and five MuMax$^3$ repeats (330 anchor--target pairs), 0.1590\%
[0.1131\%, 0.2135\%] of valid cells meet this criterion and 327/330 pairs
contain at least one such cell. The corresponding fractions are 0.0536\%
for driven transitions (162/165 pairs) and 0.2645\% after relaxation (165/165
pairs). The interval resamples 33 base groups while retaining both segment
roles and all five repeats. Thus the fallback is rare at cell level but not
an ignorable corner case at transition level.

The conditional flow bridge is Euclidean in rotation coordinates:
\begin{equation}
\begin{aligned}
\Omega_\tau&=(1-\tau)\Omega_0+\tau\Omega^\star,\\
\mathcal L_{\mathrm{FLARE}}
&=\mathbb E_{\tau,\Omega_0}
  \Bigl[\bigl\|D\odot
  \bigl(v_\theta(\Omega_\tau,\tau,c)\\
  &\qquad -(\Omega^\star-\Omega_0)\bigr)\bigr\|_2^2\Bigr],
\end{aligned}
\label{eq:app_cfm_compact}
\end{equation}
where $D$ is the magnetic mask, $\tau$ is flow time, and $c$ contains the
requested physical horizon and physical controls.  Flow time and physical
time have separate embeddings.

The source is an isotropic Gaussian over ambient rotation coordinates, as specified in the
main text.

\subsection{Architecture, optimization, and sampling}

The velocity model is a native-resolution U-Net.  Its image input concatenates
the three anchor channels, three rotation-path channels, and thirteen spatial
condition channels.  A separate global encoder embeds duration, flow time,
material parameters, field/current controls, thermal schedule, boundary
flags, geometry descriptors, and an $8\times8$ current-layout summary.  The
global embedding modulates all four U-Net resolutions.  One attention block is
used at the bottleneck.

Table~\ref{tab:app_configuration} collects the implementation and sampling
configuration used by the reported checkpoint.
\begin{table*}[!tp]
\centering
\small
\caption{Compact implementation specification.  Batch size is global across
two training devices.  The reported checkpoint is the final EMA state rather
than a test-selected checkpoint.}
\label{tab:app_configuration}
\begin{tabular}{ll@{\qquad}ll}
\toprule
Item & Value & Item & Value \\
\midrule
Grid & $256\times256$ & Trainable parameters & 65,260,803 \\
Image channels & 19 & Base width / multipliers & 64 / $[1,2,4,8]$ \\
Residual blocks & 2 per level & Bottleneck attention & 8 heads \\
Global condition width & 512 & Group normalization & 8 groups \\
Stage--1 updates & 50,000 & Global batch & 128 \\
Optimizer & AdamW & Learning rate & $2{\times}10^{-4}\rightarrow2{\times}10^{-6}$ \\
Warmup / weight decay & 2,500 / 0.01 & Adam betas & $(0.9,0.95)$ \\
Gradient clipping / EMA & 1.0 / 0.9999 & Arithmetic & bfloat16 AMP \\
Training seed / split seed & 78 / 1234 & Augmentation & in-plane $90^\circ$ rotations \\
Primary sampler & Heun--10 & Network evaluations & 20 per segment \\
\bottomrule
\end{tabular}
\end{table*}

Stage~1 is trained from random initialization.  Stage~2 begins from the
Stage~1 EMA and receives 25,000 additional updates at learning rate
$10^{-5}$.  For later segments, two unfiltered Stage~1 rollout states are generated
in advance with the same Heun--10 sampler used for evaluation.  The primary
Stage~2 model draws exact and generated starts with probability $1/2$ each.
The GT-only and generated-state-only controls use the same initialization, update
budget, and cached rollout states, changing only the mixture weights.  Whenever the
anchor is replaced, the target rotation is recomputed relative to
that anchor.

Heun--10 denotes ten equal flow-time steps and therefore twenty velocity
evaluations per physical segment, or 40 network evaluations for a complete two-segment output.  A fresh source field is drawn for each physical segment.  During
autoregressive rollout, only the magnetic state is propagated: the material
and future control schedule remain externally specified.  Thus the rollout
test measures state-rollout error without introducing control-policy error.

\subsection{Data partition and evaluation units}

The x5 corpus contains finite-temperature MuMax$^3$ trajectories with driven and
post-drive relaxation segments.  Table~\ref{tab:app_dataset_definition}
reports the realized support after generation; min--max entries are measured
over all completed records rather than copied from proposal priors.  In
addition to the x5 training/evaluation corpus, the simulator-only audit uses
an x30 collection generated under the same physical protocol with 30 distinct
thermal seeds for each selected base condition.  In particular, ``disk'',
``ellipse'', ``ring'', ``stripe cut'', and ``two lobe''
refer to the spatial \emph{current-mask} layout on an $8\times8$ control grid.
The magnetic body itself is the same square film in every run, so the ring
shift studied below is a held-out drive-mask family, not a change from a square
film to an annular sample.

\begin{table*}[!tp]
\centering
\small
\caption{Definition and realized support of the custom x5 MuMax$^3$ corpus.
Counts in parentheses are numbers of the 1,000 base conditions; stochastic
thermal repeats inherit their base parameters. Fixed material entries are
singletons, not omitted ranges.}
\label{tab:app_dataset_definition}
\begin{tabular}{p{0.23\textwidth}p{0.70\textwidth}}
\toprule
Quantity & Value or realized support \\
\midrule
Film and discretization & $256\times256\times1$ cells of
$2\times2\times1$~nm$^3$; $512\times512\times1$~nm square film; open
boundaries; edge smoothing 8 \\
Material constants & $M_s=5.8\times10^5$~A\,m$^{-1}$,
$A_{\rm ex}=1.5\times10^{-11}$~J\,m$^{-1}$,
$K_{u1}=8.0\times10^5$~J\,m$^{-3}$, and $\alpha=0.30$ in every base \\
DMI content & Interfacial Dzyaloshinskii--Moriya interaction in every base,
$D_{\rm ind}=3.25\times10^{-3}$~J\,m$^{-2}$; no zero-DMI or bulk-DMI cases \\
Temperature & $30$~K (334), $150$~K (333), or $300$~K (333), with an
independent MuMax$^3$ thermal seed for each repeat \\
Applied field & Uniform $B_z\in\{0,8,16,24,32\}$~mT
(200/200/200/201/199) \\
SOT current & Slonczewski SOT on a static $8\times8$ spatial mask;
signed nonzero drive amplitude $-4.487\times10^{12}$ to
$4.413\times10^{12}$~A\,m$^{-2}$, with
$|J|\in[0.404,4.487]\times10^{12}$~A\,m$^{-2}$; $J=0$ after the pulse \\
Current-mask families & Disk (204), ellipse (200), ring (209), stripe cut
(196), and two lobe (191); centers span approximately $[-170,170]$~nm in
both axes, with radius 14.0--54.0~nm, width 7.0--22.0~nm, and aspect ratio
0.55--1.80 where applicable \\
Initial textures & Helical (505) or stripe (495); sampled helical period
54.0--86.0~nm and wall width 5.01--9.00~nm \\
Temporal protocol & 1.0~ns shared initial relaxation, 0.202--0.998~ns
current pulse, and 3.5~ns zero-current relaxation \\
Corpus and split & 7,770 completed trajectories, 1,000 base conditions;
6,227/786/757 train/validation/test trajectories under split seed 1234 \\
\bottomrule
\end{tabular}
\end{table*}

The indivisible split unit is a
\emph{base condition}: all stochastic repeats, saved frames, and control
segments sharing that physical setup remain in one partition.  This prevents
near-duplicate frames or repeat trajectories from crossing the split.
Training statistics are fit on the training partition only.  The configured
1--6\,ns filter applies to \emph{absolute frame times}.  Legal training pairs
remain inside a constant-control segment: the fixed duration grid contributes
0.25, 0.5, 1, 2, and 3\,ns pairs where feasible, and the longest realized
single segment is the 3.5\,ns post-drive relaxation.  Consequently there are
no legal one-shot 4--6\,ns targets in this corpus; a control discontinuity or
longer protocol is handled by explicit segment composition.

Table~\ref{tab:app_evaluation_units} fixes the data, draw, and clustering units
used by each frozen evaluation.
\begin{table*}[!tp]
\centering
\scriptsize
\caption{Data and statistical units used by the frozen evaluation.  Pixels,
draws, repeats, and segments are never treated as independent
confidence-interval samples.}
\label{tab:app_evaluation_units}
\setlength{\tabcolsep}{3pt}
\begin{tabular}{@{}p{.19\textwidth}p{.10\textwidth}p{.08\textwidth}p{.08\textwidth}p{.23\textwidth}p{.25\textwidth}@{}}
\toprule
Collection & Items & Base clusters & Draws per item & Segment role & Use \\
\midrule
Training / validation / test trajectories & 6,227 / 786 / 757 & split by base & -- & both &
Optimization and held-out evaluation \\
Single segment & 1,493 cases & 100 & 4 & within segment &
Single-segment evaluation \\
Complete-path prediction distribution & 33 conditions & 33 & 5 & drive $\rightarrow$ predicted handoff $\rightarrow$ relax &

Distribution Score and angular energy distance \\

MuMax$^3$ repeat reference & 34 conditions & 17 & 30 & 17 drive + 17 relax &
Two-repeat angle, five-versus-five distances, and driven exact-anchor score \\
Exact-anchor complete paths (Table~\ref{tab:main_distribution}) & 165 anchors & 33 & 5 stochastic / 1 deterministic & drive $\rightarrow$ predicted handoff $\rightarrow$ relax &
Fair energy score \\

Multi-stage rollouts & 736 rollouts & 97 & fixed & two or more &
Teacher-forced and autoregressive evaluation \\
Semigroup test & 512 cases & 69 & ensembles retained & adjacent spans &
One-shot versus composition \\
Ring OOD, single segment / rollout & 3,126 / 1,518 OOD rows & 209 / 198 & fixed & both &
Held-out geometry family \\
\bottomrule
\end{tabular}
\end{table*}

Validation and test anchors are deterministic within each legal interval.
Training uses random anchors and in-plane quarter-turn augmentation, with
vector components and spatial controls transformed together.  Of the generic
full-resolution physical channels, the geometry mask and $j_z$ carry spatial
information in x5; material, field, and temperature variation is supplied
through the global condition path.

\subsection{Comparator and timing protocol}
All learned-dynamics comparators receive 50,000 x5 adaptation updates, and each
method is evaluated using its native temporal interface.  Main quality follows
one complete exact-control path: the saved pre-drive state is advanced through
the realized driven segment, the model prediction is handed to the exact
3.5-ns zero-current relaxation, and only the final post-relaxation endpoint is
scored.  FLARE, Cartesian CFM, Poseidon-T, and CNO-FM make one exact-duration
endpoint query per segment; DPOT-Ti, MPP-AViT-Ti, and PDEArena use causal
0.25-ns advances plus a fractional tail interval; LE-PDE re-encodes at the
handoff and uses the nearest enclosing integer number of latent advances; and
NeuralMAG-x5 integrates each exact segment with its LLG/RK4 loop.  No saved MuMax$^3$ boundary state is required or supplied at the handoff.
Because these physical quality paths have different durations, speed is
measured on a separate standardized fixed two-segment 5-ns compute workload:
a 2-ns driven segment followed by a 3-ns zero-current relaxation, with one
predicted-state handoff. On this workload, Poseidon-T and CNO-FM make one
lead-time query per segment (two total); DPOT-Ti, MPP-AViT-Ti, and PDEArena use
eight plus twelve causal 0.25-ns steps with continuous autoregressive history;
LE-PDE re-encodes at the handoff and performs two encodes, twenty latent
advances, and two decodes; FLARE uses ten Heun steps per segment (twenty steps
and forty velocity evaluations total); and NeuralMAG-x5 uses 20,000 plus
30,000 RK4 steps (200,000 learned-demagnetization calls total). Outputs are
mapped to the common $256^2$ metric grid. Timings are measured after warmup on
one NVIDIA H200, exclude data loading and host transfer, and cover the complete
representative workload rather than an isolated network call.
The seven author-released methods named above, together with the internal
Cartesian-CFM representation control, constitute the comparator set in
Table~\ref{tab:main_distribution}.  The FNO used in the
held-out-ring and magnetization-field audits in Appendices~\ref{app:ood}
and~\ref{app:visual} is instead a
60.55M-parameter, four-block, 20-mode deterministic internal-architecture
control trained for 50,000 x5 updates.  It is used only to check whether the
ring-mask transfer pattern and the displayed local defects are specific to
FLARE; it is not included in the primary distribution leaderboard or timing
claims.

For each learned or hybrid method, we test powers-of-two batch sizes from one
through a method-specific ceiling or the first unsupported or out-of-memory
setting. Appendix~\ref{app:solver_time} reports every supported measurement.
Table~\ref{tab:main_distribution} reports two separate ratios:
MuMax$^3$ batch-one latency divided by learned-model $B=1$ latency, and the same
MuMax$^3$ reference divided by the learned model's lowest measured per-output
latency. MuMax$^3$ retains its native batch one.
In a separate single-H200 concurrency audit, standardized representative 5-ns repetitions at
$B=2$ and 4 were respectively $2.34\times$ and $2.39\times$ slower per output
than the audit's $B=1$ run.  A fixed-step 0.2-ns sweep through $B=32$
likewise made every $B\geq2$ setting $2.11$--$2.16\times$ slower per output.
We therefore retain MuMax$^3$'s native batch one in the representative 5-ns comparison.

\paragraph{Search and checkpoint-selection budget.}

The external endpoint adapters first received one common 10,000-update bring-up
run, followed by the frozen 50,000 run reported here. Poseidon-T, CNO-FM,
DPOT-Ti, and PDEArena use the recorded AdamW recipe, LE-PDE uses Adam, and MPP
uses the released DAdaptAdan recipe after one same-stage optimizer pilot
outperformed AdamW. No architecture sweep or test-set early stopping is used
for the reported rows. FLARE and every endpoint baseline use the final 50,000
checkpoint (FLARE uses its EMA state), and the quality checkpoint is not
chosen for timing. For NeuralMAG, we report the final 50,000 component-adaptation
checkpoint rather than the excluded short pilot checkpoints; diagnostic evaluations during
adaptation are not used for early stopping or checkpoint selection. Thus the
search asymmetry is explicit: one optimizer pilot benefits MPP, while the
other primary recipes are fixed by the standardized or author-native
protocol.

\section{Metrics and uncertainty}
\label{app:metrics}

\subsection{Field, topology, and ensemble metrics}

For valid cells $\mathcal V$, the pointwise field metrics are
\begin{equation}
\begin{aligned}
e_{\angle}(\hat m,m)
&=\frac{1}{|\mathcal V|}\sum_{i\in\mathcal V}
\frac{180}{\pi}\arccos\!\left(
\operatorname{clip}(\hat m_i^\top m_i,-1,1)\right),\\
e_{\mathrm{MSE}}
&=\frac{1}{|\mathcal V|}\sum_{i\in\mathcal V}
\|\hat m_i-m_i\|_2^2.
\end{aligned}
\label{eq:app_field_metrics}
\end{equation}
Here, MSE is the mean squared vector error per valid cell: the squared
errors of the three magnetization components are summed before averaging
over cells. This is the convention used in Table~\ref{tab:design_ablation}(b).
We additionally report absolute topological-charge error and the absolute
difference in a dimensionless energy-density diagnostic.  These global
observables are complementary to field angle: two states can have similar
energy yet place domain walls differently, while a small average field error
can still conceal a topology-changing local defect.

Distribution fidelity uses five MuMax$^3$ repeats and five model predictions for
each macro-condition.  The local-translation-matching distance searches over
a bounded displacement window before comparing patches, making it tolerant
to physically equivalent small translations of a skyrmion or wall.  The
Distribution Score is defined in Section~\ref{sec:distribution_metric} of the main paper.

For $X=\{x_i\}_{i=1}^{M}$ and $Y=\{y_j\}_{j=1}^{N}$, its kernel and
MuMax-only bandwidth are
\begin{equation}
\begin{gathered}
K_\sigma(d)=\exp\!\left(-\frac{d^2}{2\sigma^2}\right),\\
\sigma=\max\!\left\{10^{-6},\;
\operatorname*{median}_{i}\min_{k\ne i}D(x_i,x_k)\right\},
\end{gathered}
\label{eq:app_distribution_kernel}
\end{equation}
and the two densities evaluated at each MuMax$^3$ repeat are
\begin{equation}
\begin{gathered}
q_{\mathrm{model}}(x_i)=\frac{1}{N}\sum_{j=1}^{N}
K_\sigma\!\left(D(x_i,y_j)\right),\\
q_{\mathrm{MuMax}}^{-i}(x_i)=\frac{1}{M-1}\sum_{k\ne i}
K_\sigma\!\left(D(x_i,x_k)\right).
\end{gathered}
\label{eq:app_distribution_densities}
\end{equation}
Thus the bandwidth is fixed before inspecting model outputs and the MuMax$^3$
density is leave-one-out.

Scores are
aggregated geometrically over conditions, so a near-zero failure on one
physical regime cannot be hidden by many easy cases.  Angular energy distance
supplies a second, unordered two-sample comparison:
\begin{equation}
\begin{aligned}
\widehat{\mathrm{ED}}_\angle
&=\frac{2}{LM}\sum_{\ell,j}d_\angle(x_\ell,y_j)
\\ &\quad -\frac{1}{L^2}\sum_{\ell,\ell'}d_\angle(x_\ell,x_{\ell'})
\\ &\quad -\frac{1}{M^2}\sum_{j,j'}d_\angle(y_j,y_{j'}).
\end{aligned}
\label{eq:app_energy_distance}
\end{equation}
Repeat-paired angular error is also retained because it answers a different
question: accuracy for a particular stochastic realization rather than
coverage of the unordered outcome set.

For an exact anchor with observed target $x$ and $L=5$ stochastic forecasts
$y_1,\ldots,y_L$, the finite-ensemble-corrected fair energy score is
\begin{equation}
\widehat{\mathrm{ES}}_{\mathrm{fair}}
=\frac{1}{L}\sum_{\ell=1}^{L}d_{\mathrm{chord}}(y_\ell,x)
-\frac{1}{L(L-1)}\sum_{\ell<\ell'}d_{\mathrm{chord}}(y_\ell,y_{\ell'}),
\label{eq:app_fair_energy_score}
\end{equation}
where $d_{\mathrm{chord}}$ is the RMS Euclidean chord distance over all valid-cell unit vectors. Deterministic single checkpoints are evaluated as one-point distributions. Unlike angular energy distance, this is a proper score against one observed target rather than an unordered two-sample distance.

For the simulator-only reference, the 30 MuMax$^3$ repeats of each x30 condition
are deterministically partitioned into six disjoint folds of five.  We average
angular energy distance and Distribution Score over all 30 ordered pairs of
distinct folds; the two-repeat angular reference averages every ordered pair
of distinct trajectories.  The driven branches all load the same relaxed
checkpoint before applying independent thermal seeds.  This permits a proper
exact-anchor MuMax$^3$ fair-energy reference: one five-repeat fold is treated as
the forecast ensemble and endpoints in disjoint folds are observations.  The
post-relaxation starts are already branch-specific, so we do not report a
post-relaxation or combined exact-anchor fair-energy reference.  The fair
energy reference is positive even for an oracle distribution because it
retains the physical outcome spread; ``noise floor'' here means the
simulator-only reference level, not an algebraic zero.

\subsection{Clustered intervals}

All 95\% intervals resample base conditions.  Every repeat, segment, draw,
and pixel belonging to a sampled base condition remains together.  Paired
ablations resample the same base indices and compute within-case differences
before aggregation.  The intervals therefore describe variation across
held-out physical conditions for fixed trained checkpoints.
They do not include simulator-parameter uncertainty or model-family selection uncertainty; training-seed variation is reported separately in Appendix~\ref{app:seed_geometry}.

\section{Distribution fidelity}
\label{app:distribution}
\begin{figure*}[!tp]
\centering
\includegraphics[width=\textwidth]{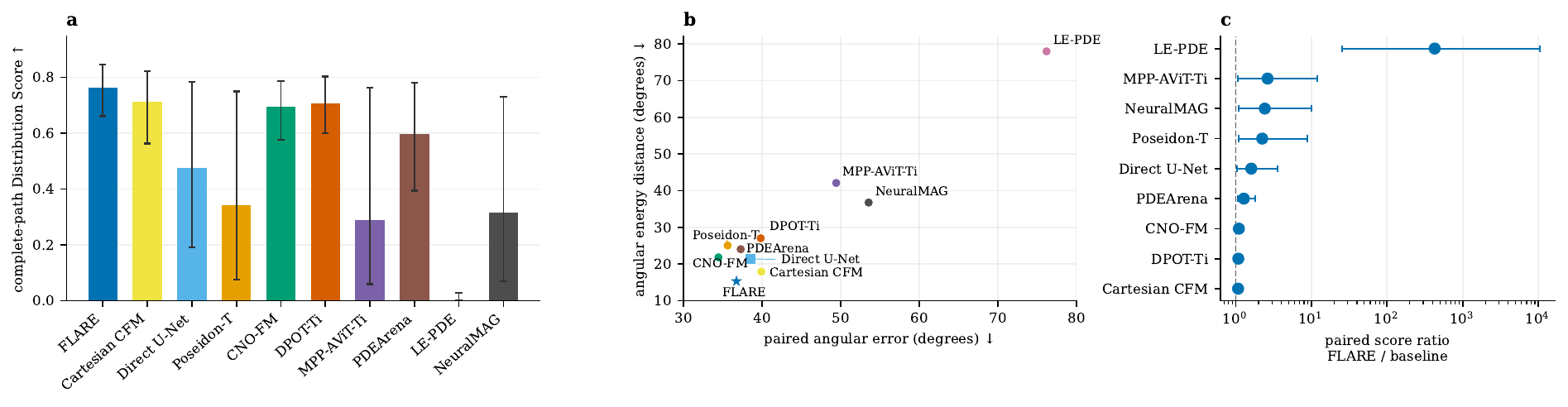}

\caption{\textbf{Exact-control complete-path distribution diagnostics.}
\textbf{(a)} Distribution Score at the final post-relaxation endpoint after
prediction-only handoff.  \textbf{(b)} Repeat-paired angle versus angular
energy distance; lower-left is better and the star marks FLARE.  The two
axes need not rank a stochastic model identically.  \textbf{(c)} Paired
Distribution-Score ratios for FLARE relative to each 50,000 learned-dynamics
baseline with a reported Distribution Score; ratios above one favor FLARE
and bars are 95\% base-cluster intervals.  The plotted comparison uses five
outcomes for every method and 33 complete-path conditions from Table~\ref{tab:main_distribution}.}

\label{fig:app_distribution_diagnostics}
\end{figure*}

Figure~\ref{fig:app_distribution_diagnostics} separates complete-path
distribution fidelity, paired accuracy, and paired score ratios.
In Table~\ref{tab:main_distribution} (seed 78), FLARE obtains a Distribution Score of
0.762 [0.660, 0.844] and an angular energy distance of $15.30^\circ$ [13.32,
17.26].  Cartesian CFM is the matched representation control at 0.711; Direct U-Net,
the direct-regression control, scores 0.477. Paired bootstrap gives a 7.14\%
[0.00\%, 19.46\%] FLARE gain over Cartesian CFM.  FLARE has the best reported
Distribution Score and angular energy distance.  This is
an expected distinction between matching one indexed realization and covering
a stochastic outcome set, so the metrics are not collapsed into one weighted
leaderboard.

\subsection{Conditional outcome variability under identical physical inputs}
\label{app:generative_necessity}

Finite-temperature micromagnetic evolution exhibits substantial stochastic
variability even when the exact magnetization anchor, physical controls, and
prediction horizon are held fixed. To examine whether FLARE captures this
conditional outcome distribution, we compare 30 independent MuMax$^3$ thermal
realizations with 30 fresh Stage~1 FLARE ODE--10 draws and the single output of
the deterministic Poseidon-T baseline under identical inputs. All MuMax$^3$
repeats within each case share the same exact anchor and control schedule.

\begin{figure*}[!tp]
    \centering
    \includegraphics[width=\textwidth]{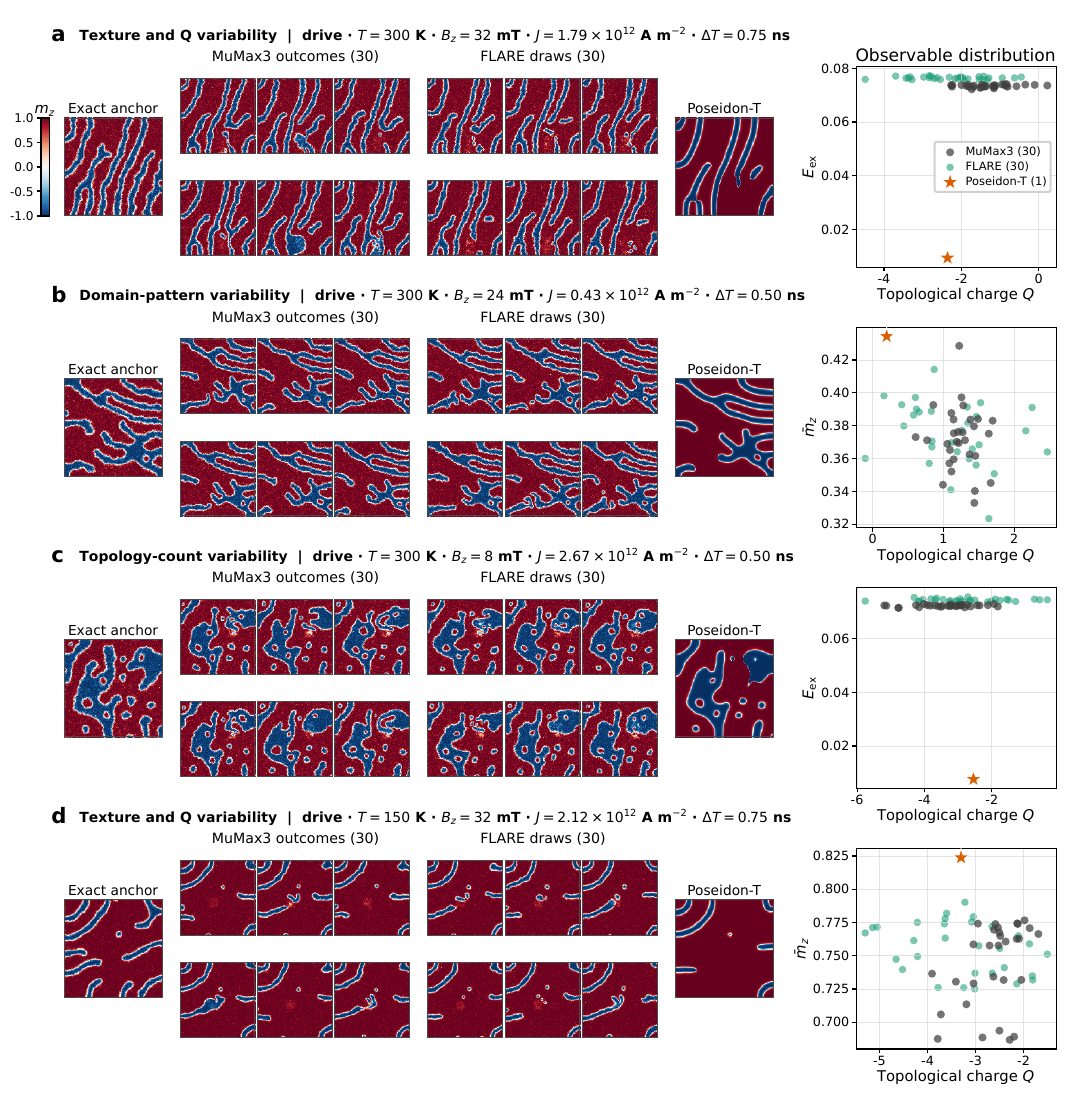}
    \caption{\textbf{Conditional outcome variability under identical physical
    inputs.}
    Each row fixes the exact magnetization anchor, physical controls, and target
    horizon, and compares 30 independent MuMax$^3$ thermal outcomes, 30 fresh
    Stage~1 FLARE ODE--10 draws, and one deterministic Poseidon-T prediction.
    Six representative fields from each stochastic ensemble are selected using
    the same deterministic medoid-based rule, while all 30 outcomes are included
    in the observable scatter plot.
    \textbf{(a)} Texture and topological-charge variability.
    \textbf{(b)} Domain-pattern variability.
    \textbf{(c)} Topology-count variability.
    \textbf{(d)} An additional case with texture and topological-charge
    variability.
    Across these cases, FLARE generates diverse future states that reproduce
    substantial portions of the MuMax$^3$ outcome spread in both field space and
    observable space, whereas the deterministic baseline provides only a single
    prediction for each fixed input.}
    \label{fig:app_generative_necessity}
\end{figure*}

Figure~\ref{fig:app_generative_necessity} shows that FLARE captures multiple
forms of finite-temperature outcome variability. In some conditions, MuMax$^3$
realizations exhibit continuous variation in domain patterns while retaining
similar global topology; in others, the stochastic evolution produces a broad
range of topological-charge outcomes. FLARE generates corresponding ensembles
with substantial overlap with the MuMax$^3$ outcome clouds and reproduces both
texture-level and observable-level variation across the selected cases.
In contrast, a deterministic predictor maps each fixed input to only one future
state. These results directly illustrate the advantage of modeling the
conditional future state as a distribution: FLARE can represent the physically
plausible spread of stochastic micromagnetic outcomes rather than reducing the
prediction to a single endpoint.

\subsection{Sensitivity, simulator self-consistency, and physical observables}
\label{app:distribution_observables}
Table~\ref{tab:app_distribution_sensitivity} reports the kernel and
translation-matching sensitivity checks from the separate frozen
single-segment evaluation.
\begin{table*}[!tp]
\centering
\small
\caption{Single-segment Distribution Score sensitivity. The bandwidth multiplier rescales only the MuMax$^3$ nearest-neighbor bandwidth; ``no translation'' removes local translation matching while retaining the nominal bandwidth rule. The last column is a paired base-group bootstrap ratio; these values are not the primary complete-path results in Table~\ref{tab:main_distribution}.}
\label{tab:app_distribution_sensitivity}
\begin{tabular}{lccc}
\toprule
Setting & FLARE & Best comparator & FLARE/comparator [95\% CI] \\
\midrule
Bandwidth $0.5\times$ & 0.3274 & PDEArena 0.2330 & 1.405 [1.256, 1.590] \\
Bandwidth $1\times$ & 0.8050 & PDEArena 0.7362 & 1.094 [1.059, 1.132] \\
Bandwidth $2\times$ & 0.9543 & PDEArena 0.9287 & 1.028 [1.016, 1.040] \\
No translation & 0.8318 & PDEArena 0.7769 & 1.071 [1.039, 1.107] \\
\bottomrule
\end{tabular}
\end{table*}

Within this single-segment diagnostic, FLARE ranks first under every tested bandwidth and without translation matching, so its ordering is not an artifact of the nominal kernel choice.
Table~\ref{tab:app_mumax_reference} gives the independent simulator-only
reference levels used for absolute interpretation.
\begin{table*}[!tp]
\centering
\scriptsize
\setlength{\tabcolsep}{3.2pt}
\caption{Independent MuMax$^3$--MuMax$^3$ reference levels for absolute
interpretation of the distribution metrics.  The audit uses 30 thermal
repeats for each of 17 held-out base conditions; intervals resample the 17
bases and retain both segment roles.  Distinct-fold rows average all ordered
pairs of disjoint five-repeat folds.  These are simulator-only finite-sample
references, not algebraic zeros or paired estimates on the 33 main bases.}
\label{tab:app_mumax_reference}
\resizebox{\textwidth}{!}{%
\begin{tabular}{lcccc}
\toprule
Metric & Combined [95\% CI] & Driven [95\% CI] & Post-relaxation [95\% CI] & MuMax$^3$ repeat protocol \\
\midrule
Distribution Score $\uparrow$ & 0.901 [0.866, 0.930] & 0.941 [0.927, 0.953] & 0.862 [0.801, 0.915] & Disjoint five versus five \\
Two-repeat angle ($^\circ$) $\downarrow$ & 28.86 [22.84, 34.82] & 27.47 [21.51, 33.37] & 30.25 [23.89, 36.66] & All distinct trajectory pairs \\
Angular energy distance ($^\circ$) $\downarrow$ & 11.60 [9.17, 14.07] & 10.99 [8.62, 13.35] & 12.20 [9.62, 14.86] & Disjoint five versus five \\
Fair energy score $\downarrow$ & -- & 0.283 [0.230, 0.335] & -- & Same exact anchor; disjoint forecasts/targets \\
\bottomrule
\end{tabular}}
\end{table*}
The Table~\ref{tab:main_distribution} complete-path point estimates are above
the independent final post-relaxation scales: FLARE's paired-angle excess is
$6.47^\circ$, and its angular-energy excess is $3.10^\circ$.  These are
descriptive absolute-scale comparisons rather than lower-bound gaps because
the main model and target paths share a drive-start anchor whereas independent
MuMax$^3$ pairs do not.
For the segment in which an exact-anchor oracle is available, FLARE's driven
fair energy score is 0.322 [0.290, 0.353], versus the MuMax$^3$ reference 0.283
[0.230, 0.335].  This directly distinguishes errors of the order of
finite-temperature repeat variability from much larger discrepancies.  It
does not erase the subset limitation: the reference uses a separate 17-base
x30 audit, whereas the main comparison uses 33 x5 bases, so we do not treat
the differences as paired hypothesis tests or the simulator values as
universal bounds.  A same-five calculation that reuses a trajectory on both
sides is excluded because it would produce an optimistically biased null.

\subsection{Metric and finite-ensemble pressure tests}
\label{app:metric_pressure}
\begin{figure*}[!tp]
\centering
\includegraphics[width=\textwidth]{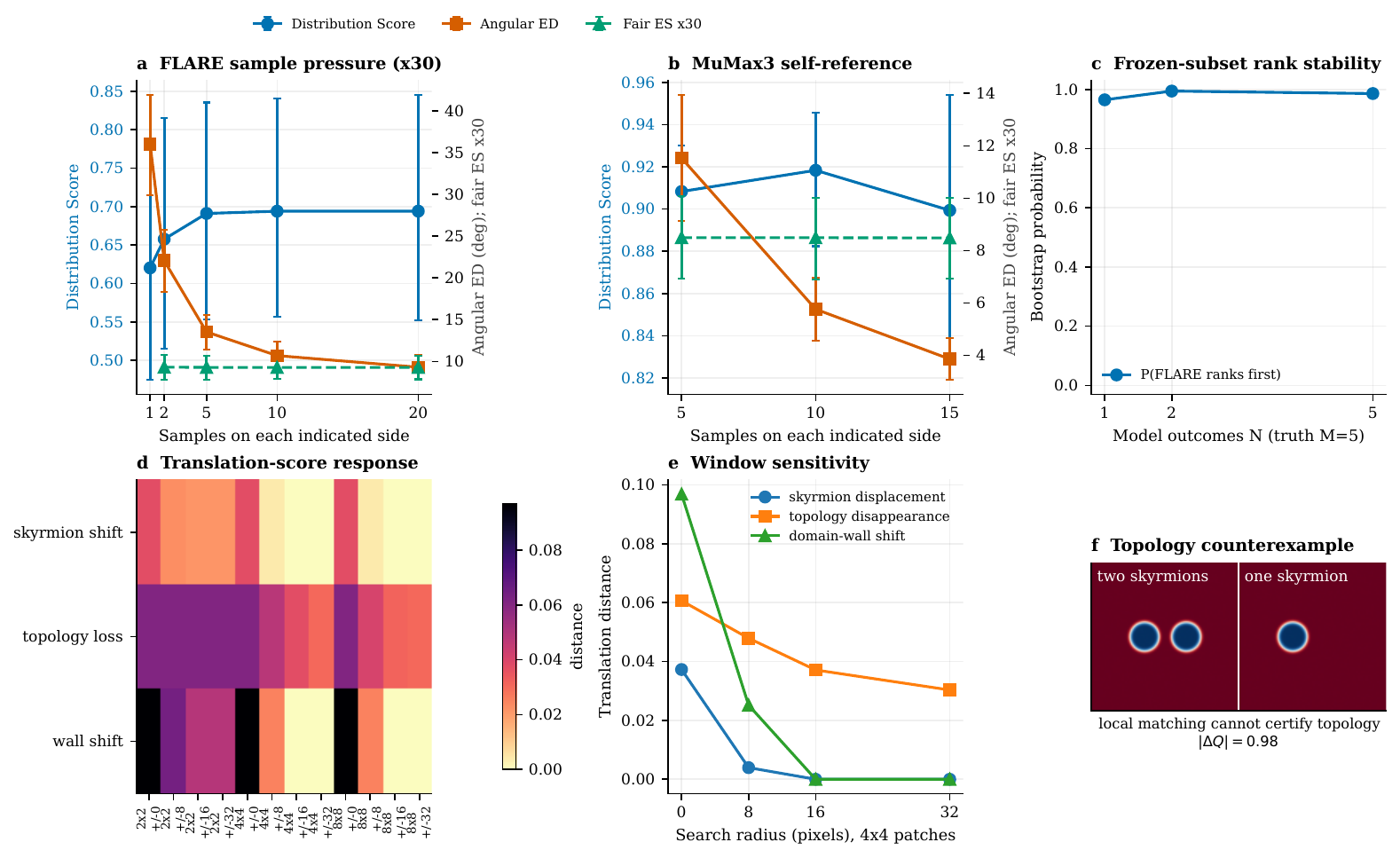}
\caption{\textbf{Metric, sample-budget, and translation-matching stress
tests.} \textbf{(a)} The frozen FLARE checkpoint on the independent x30
audit, using five MuMax$^3$ targets and $N\in\{1,2,5,10,20\}$ model outcomes.
\textbf{(b)} Disjoint MuMax$^3$ sets at 5-versus-5, 10-versus-10, and
15-versus-15. Fair energy score is multiplied by 30 only for display on the
angular-distance axis. \textbf{(c)} Probability that FLARE ranks first in
the frozen eight-method single-segment subset under base-cluster resampling.
\textbf{(d)} The
complete $3\times4$ patch-grid/window response for three analytic
sphere-valued perturbations. \textbf{(e)} The nominal $4\times4$ grid as a
function of search radius. \textbf{(f)} A topology-changing counterexample:
local matching can reduce the field discrepancy but cannot certify $Q$. Bars
and bands are 95\% base-cluster intervals.}
\label{fig:app_metric_stress}
\end{figure*}

Figure~\ref{fig:app_metric_stress} summarizes the sample-budget, estimator,
translation-window, and topology stress tests.

The x30 audit provides enough independent repeats to vary both sides of an
empirical distribution comparison without replacement. For each of its 17
held-out bases and two segment roles, we form 32 deterministic nested
subsampling trials. The model stress test draws from a persisted pool of 40
out of 128 frozen-checkpoint forecasts and retains five MuMax$^3$ targets; the
simulator-only test uses disjoint MuMax$^3$ subsets. Intervals resample bases
and keep both segment roles and every within-condition trial together.
Table~\ref{tab:app_metric_budget_stress} lists the corresponding numerical
sample-budget results.
\begin{table*}[!tp]
\centering
\scriptsize
\setlength{\tabcolsep}{4.0pt}
\caption{Finite-ensemble pressure test on the independent x30 audit. The
FLARE rows use $N$ model outcomes against five MuMax$^3$ outcomes; the reference
rows use two disjoint MuMax$^3$ sets of the indicated size. Fair energy score is
reported only for driven branches, where independent thermal outcomes share
the exact anchor. At $N=1$ the finite-ensemble correction is undefined; the
corresponding ordinary one-member energy score is 0.592 [0.498, 0.679].}
\label{tab:app_metric_budget_stress}
\begin{tabular}{llccc}
\toprule
Source comparison & Budget & Distribution Score $\uparrow$ & Angular energy dist. ($^\circ$) $\downarrow$ & Fair energy score $\downarrow$ \\
\midrule
FLARE--MuMax$^3$ & $1$ vs. $5$ & 0.620 [0.474, 0.781] & 36.03 [29.85, 41.91] & -- \\
FLARE--MuMax$^3$ & $2$ vs. $5$ & 0.658 [0.516, 0.816] & 22.06 [18.37, 25.74] & 0.3105 [0.2592, 0.3594] \\
FLARE--MuMax$^3$ & $5$ vs. $5$ & 0.691 [0.553, 0.836] & 13.53 [11.42, 15.60] & 0.3095 [0.2615, 0.3565] \\
FLARE--MuMax$^3$ & $10$ vs. $5$ & 0.694 [0.556, 0.841] & 10.70 [9.08, 12.37] & 0.3094 [0.2631, 0.3577] \\
FLARE--MuMax$^3$ & $20$ vs. $5$ & 0.694 [0.553, 0.845] & 9.31 [7.87, 10.75] & 0.3092 [0.2620, 0.3562] \\
\midrule
MuMax$^3$--MuMax$^3$ & $5$ vs. $5$ & 0.908 [0.886, 0.930] & 11.53 [9.15, 13.95] & 0.2831 [0.2306, 0.3357] \\
MuMax$^3$--MuMax$^3$ & $10$ vs. $10$ & 0.918 [0.882, 0.946] & 5.76 [4.56, 6.95] & 0.2831 [0.2302, 0.3341] \\
MuMax$^3$--MuMax$^3$ & $15$ vs. $15$ & 0.899 [0.819, 0.954] & 3.86 [3.06, 4.67] & 0.2827 [0.2306, 0.3340] \\
\bottomrule
\end{tabular}
\end{table*}

Two different finite-sample effects are visible. Distribution Score and the
proper fair energy score are essentially stable by five model outcomes:
increasing $N$ from 5 to 20 changes them from 0.691 to 0.694 and from 0.3095
to 0.3092. In contrast, the reported angular energy distance is the matched
biased V-statistic with zero self-distances. Its downward drift with $N$ is
an estimator effect, not evidence that the fixed model improves when sampled
more often; the same $11.53^\circ$ to $3.86^\circ$ drift appears between two
MuMax$^3$ ensembles. We therefore retain equal five-versus-five budgets for
cross-method angular-energy comparisons.
The frozen single-segment subset ordering is also stable before reaching five outcomes. At
$N=1,2,5$, FLARE has Distribution Scores 0.661 [0.426, 0.774], 0.761
[0.669, 0.824], and 0.805 [0.745, 0.857], respectively. Its median bootstrap
rank is one in every case, and the probabilities of rank one are 96.5\%,
99.5\%, and 98.6\%. Ranks above five outcomes are not extrapolated because
most primary comparators provide only five distinct
anchor-matched outputs.
Table~\ref{tab:app_translation_stress} reports the nominal-grid response to
the three analytic perturbations.
\begin{table*}[!tp]
\centering
\scriptsize
\setlength{\tabcolsep}{3.6pt}
\caption{Response of the nominal $4\times4$ local patch grid to analytic
perturbations. $D_{w}$ is the translation-matched distance with a
$\pm w$-pixel search. Figure~\ref{fig:app_metric_stress}d reports all
$2\times2$, $4\times4$, and $8\times8$ grids. Pixel MSE averages over both
cells and the three magnetization components, and therefore equals
$e_{\mathrm{MSE}}/3$ in Eq.~\ref{eq:app_field_metrics}.}
\label{tab:app_translation_stress}
\begin{tabular}{lrrrrrrr}
\toprule
Perturbation & Pixel angle & Pixel MSE & $|\Delta Q|$ & $D_0$ & $D_8$ & $D_{16}$ & $D_{32}$ \\
\midrule
Skyrmion displacement & $3.71^\circ$ & 0.02485 & 0.000 & 0.03727 & 0.00395 & 0.00000 & 0.00000 \\
Topology disappearance & $5.57^\circ$ & 0.04043 & 0.983 & 0.06064 & 0.04790 & 0.03712 & 0.03032 \\
Domain-wall shift & $9.84^\circ$ & 0.06456 & 0.000 & 0.09683 & 0.02517 & 0.00000 & 0.00000 \\
\bottomrule
\end{tabular}
\end{table*}

The two intended counterexamples delimit what translation matching does and
does not measure. First, translating an otherwise unchanged wall by 14
pixels produces a $9.84^\circ$ image-wide angular error, yet the $4\times4$
score becomes zero once the search radius reaches 16 pixels; the displaced
skyrmion behaves similarly and preserves $Q$. Thus raw pixel error can
penalize a physically equivalent placement. Second, removing one of two
skyrmions changes $|Q|$ by 0.983, while increasing the local search radius
reduces its field distance from 0.06064 to 0.03032. Local matching can
therefore attenuate a topology-changing defect. This is why every
translation-tolerant field score is accompanied by $Q$ and other global
observables rather than being treated as a sufficient physical metric. The
full grid also exposes a scale choice: at an eight-pixel window the skyrmion
distance is 0.02270, 0.00395, and 0.00381 for $2\times2$, $4\times4$, and
$8\times8$ grids, respectively; the nominal $4\times4$, $\pm16$ setting is
not uniquely selected by these examples.
Table~\ref{tab:app_physical_observables} gives the physical-observable
distribution errors and the near-antipodal prevalence audit.
\begin{table*}[!tp]
\centering
\scriptsize
\setlength{\tabcolsep}{4.0pt}
\caption{Single-segment physical-observable distribution errors and near-antipodal rotation statistics. Panel (a) reports equal-condition empirical Wasserstein-1 errors \citep{peyre2019ot} between five model and five MuMax$^3$ outcomes; lower is better. Panel (b) uses the implementation threshold.}
\label{tab:app_physical_observables}
\begin{tabular}{lrrr}
\toprule
\multicolumn{4}{l}{\textit{(a) Physical-observable distribution error}} \\
Method & $W_1(Q)$ & $W_1(\overline m_z)$ & $W_1(E_{\rm ex})$ \\
\midrule
\textbf{FLARE ODE-10} & 0.48 & \textbf{0.0283} & 0.00352 \\
Poseidon-T & 1.88 & 0.0458 & 0.03598 \\
CNO-FM & 1.64 & 0.0470 & 0.03703 \\
DPOT-Ti & 0.77 & 0.0796 & 0.03831 \\
MPP-AViT-Ti & 0.83 & 0.0827 & 0.03743 \\
PDEArena U-Net & 0.52 & 0.0840 & 0.03835 \\
LE-PDE & 4.19 &  0.3140 & 0.03231 \\
NeuralMAG-x5 & \textbf{0.35} & 0.1480 & \textbf{0.00293} \\
\midrule
\multicolumn{4}{l}{\textit{(b) Near-antipodal targets, $\theta>177.437^\circ$}} \\
Segment role & \multicolumn{2}{r}{Valid-cell fraction} & Pairs with any \\
\midrule
All & \multicolumn{2}{r}{0.1590\%} & 327/330 \\
\bottomrule
\end{tabular}
\end{table*}

Here $Q$ is the signed open-boundary finite-difference charge, mean $m_z$ is averaged over valid cells, and exchange-texture energy is the valid-neighbor mean of $1-m_i^\top m_j$. FLARE is best for mean $m_z$, while the physics-embedded NeuralMAG-x5 comparator is best for charge and exchange texture after its matched 50,000 component adaptation. These distributional observables complement rather than duplicate local-translation field similarity.

\section{Solver, efficiency, and target-time diagnostics}
\label{app:solver_time}
\begin{figure*}[!tp]
\centering
\includegraphics[width=\textwidth]{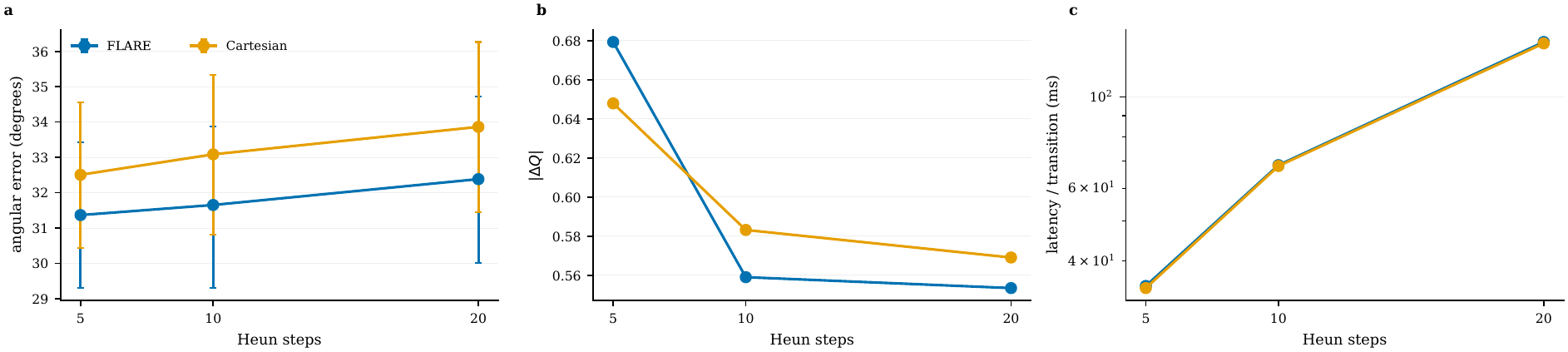}
\caption{\textbf{Heun-step sensitivity on 1,493 held-out transitions.}
\textbf{(a)} Angular error with base-cluster intervals.
\textbf{(b)} Absolute topological-charge error.
\textbf{(c)} Amortized latency per transition at batch size 32.  Increasing from 5 to 20 steps is
nearly linear in cost but does not reduce angular error, indicating that the
learned vector field, rather than integration truncation, dominates current
error.  ODE--10 is kept fixed across single-segment, cache, and rollout evaluation.}
\label{fig:app_solver_sensitivity}
\end{figure*}

\begin{figure*}[!tp]
\centering
\includegraphics[width=0.92\textwidth]{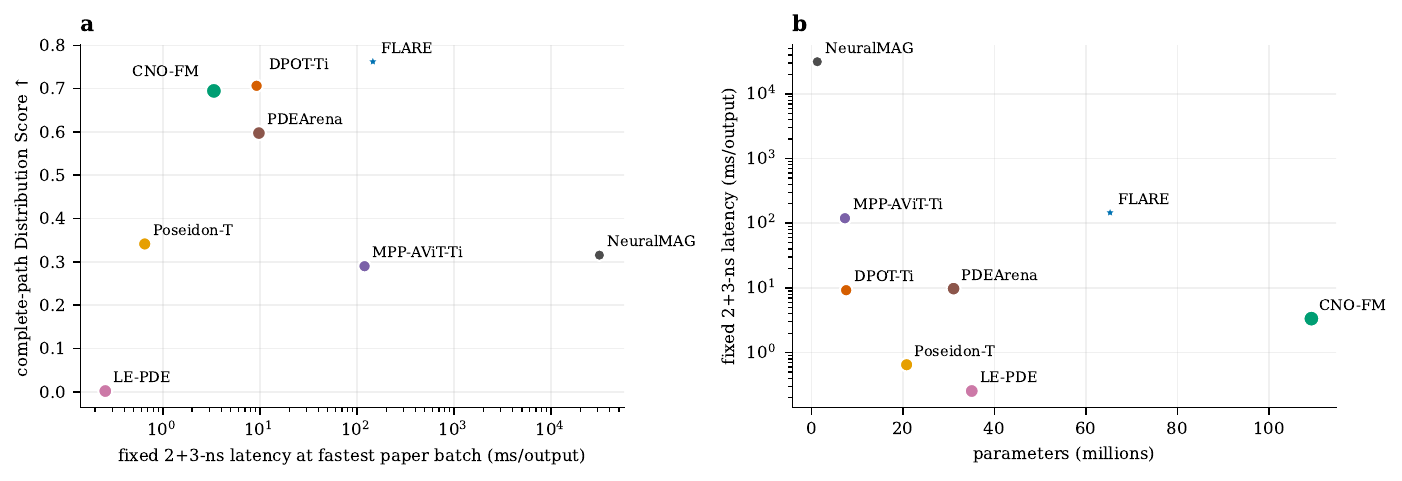}
\caption{\textbf{Complete-path quality and representative compute tradeoffs.}
\textbf{(a)} Complete-path Distribution Score versus standardized fixed two-segment
5-ns latency at each method's fastest reported batch;
marker area encodes parameter count.  \textbf{(b)} Parameter count versus
representative 5-ns latency.  Parameter count alone does not determine runtime because temporal
interfaces range from two direct-horizon predictions to forty velocity
evaluations.  FLARE remains non-dominated in the quality--throughput plane.}
\label{fig:app_compute_tradeoffs}
\end{figure*}

Figure~\ref{fig:app_solver_sensitivity} reports the Heun-step accuracy--cost
sensitivity.
For Stage~1, taking 5, 10, and 20 Heun steps gives angular errors
$31.37^\circ$, $31.65^\circ$, and $32.38^\circ$, with latencies
 34.8, 68.3, and 135.9\,ms per transition, amortized at batch size 32, in the formal sweep. Topological
error improves mildly with additional steps, so the sweep exposes a real
multi-metric tradeoff.  We retain ODE--10 because it is the sampler used to
construct the Stage~2 rollout-input distribution and was fixed for the primary
experiments, not because it is a post-hoc optimum for one test metric.
Figure~\ref{fig:app_compute_tradeoffs} compares complete-path distribution
quality, standardized fixed two-segment 5-ns latency, and parameter count.
Table~\ref{tab:app_batch_scaling} records every supported powers-of-two batch
measurement.
\begin{table*}[!tp]
\centering
\scriptsize
\setlength{\tabcolsep}{2.4pt}
\caption{Standardized fixed two-segment 5-ns (2-ns drive plus 3-ns relaxation, with one predicted-state handoff) single-H200 batch scaling. Entries are mean milliseconds per output; bold marks the lowest measured latency per method. A dash denotes an unsupported or unrecorded setting. MuMax$^3$ has no ensemble-batch interface and is shown only at its native batch one. Every learned entry uses the reported 50,000 checkpoint.}
\label{tab:app_batch_scaling}
\resizebox{\textwidth}{!}{%
\begin{tabular}{lrrrrrrrrrrrr}
\toprule
Method / batch $B$ & 1 & 2 & 4 & 8 & 16 & 32 & 64 & 128 & 256 & 512 & 1024 & 2048 \\
\midrule
FLARE ODE-10 & 995.2 & 554.6 & 353.2 & 245.8 & 191.6 & 163.9 & 150.3 & \textbf{145.7} & -- & -- & -- & -- \\
Poseidon-T & 61.91 & 32.61 & 16.90 & 8.451 & 4.140 & 2.223 & 1.268 & 0.8515 & 0.7038 & 0.6689 & \textbf{0.6499} & 0.7609 \\
CNO-FM & 72.89 & 42.73 & 21.44 & 10.76 & 5.415 & 3.824 & 3.569 & 3.429 & 3.377 & \textbf{3.354} & 3.382 & 3.406 \\
DPOT-Ti & 91.13 & 49.80 & 28.55 & 17.65 & 12.32 & 9.906 & \textbf{9.231} & 9.591 & 9.448 & -- & -- & -- \\
MPP-AViT-Ti & 613.4 & 332.4 & 194.2 & 147.1 & 135.0 & 123.7 & 121.4 & \textbf{119.6} & -- & -- & -- & -- \\
PDEArena U-Net & 40.17 & 20.53 & 14.15 & 11.58 & 10.58 & 10.12 & 9.871 & \textbf{9.758} & 9.892 & 10.61 & 10.94 & -- \\
LE-PDE & 7.642 & 3.325 & 1.764 & 0.9174 & 0.5388 & 0.3929 & 0.3158 & 0.2792 & 0.2606 & 0.2575 & 0.2560 & \textbf{0.2550} \\
NeuralMAG-x5 & 335786 & 170092 & 84963 & 45774 & 37566 & 34184 & 32402 & \textbf{31420} & -- & -- & -- & -- \\
\midrule
MuMax$^3$ & 446125 & -- & -- & -- & -- & -- & -- & -- & -- & -- & -- & -- \\
\bottomrule
\end{tabular}%
}
\end{table*}
FLARE takes 995.204\,ms at $B=1$ and reaches 145.680\,ms per output at its
fastest measured setting, $B=128$. Against the 446.125-s native-batch-one
MuMax$^3$ reference, these give a $448.3\times$ batch-one latency ratio and a
$3{,}062.4\times$ best-batch throughput ratio, respectively.
Table~\ref{tab:main_distribution} uses the same model-specific selection rule
while retaining $B=1$ as the interactive-latency column. Learned endpoint
timings average ten repeats after warmup; because one NeuralMAG-x5 fixed two-segment
5-ns workload already executes 200,000 learned-demagnetization calls, it uses one
complete timed repeat at each batch.
Table~\ref{tab:app_compute_accounting} reports training and inference compute
and memory for the final checkpoints.
\begin{table*}[!tp]
\centering
\scriptsize
\setlength{\tabcolsep}{3.0pt}
\caption{Compute and memory accounting for Table~\ref{tab:main_distribution}'s 50,000-update checkpoints. Training time is accelerator device-hours, so FLARE's two-device wall time is multiplied by two. Training peaks for the external endpoint models are PyTorch allocated-memory high-water marks; FLARE's starred value is the maximum 10-s process-memory sample per device and is therefore a more inclusive measurement. Inference peaks use the fixed two-segment representative 5-ns $B=1$ query; the final column shows the memory cost at each method's fastest measured batch.}
\label{tab:app_compute_accounting}
\resizebox{\textwidth}{!}{%
\begin{tabular}{lp{0.30\textwidth}rrrr}
\toprule
Method & Work in one fixed 2+3-ns timed output & Train H200-h & Train peak (GiB) & $B=1$ peak (GiB) & Fastest $B$ / peak (GiB) \\
\midrule
FLARE ODE-10 & 20 Heun steps; 40 velocity evaluations & 15.20 & $96.9^\ast$ & 0.539 & 128 / 18.54 \\
Poseidon-T & Two lead-time-conditioned endpoint calls & 3.71 & 1.15 & 0.182 & 1024 / 65.62 \\
CNO-FM & Two lead-time-conditioned endpoint calls & 6.83 & 4.61 & 0.493 & 512 / 27.44 \\
DPOT-Ti & 20 recurrent 0.25-ns calls & 6.91 & 2.38 & 0.350 & 64 / 18.61 \\
MPP-AViT-Ti & 20 recurrent 0.25-ns calls & 6.77 & 12.29 & 0.887 & 128 / 105.97 \\
PDEArena U-Net & 20 recurrent 0.25-ns calls & 0.953 & 1.58 & 0.270 & 128 / 12.55 \\
LE-PDE & Two encodes, 20 latent advances, two decodes & 1.72 & 1.15 & 0.277 & 2048 / 101.23 \\
NeuralMAG-x5 & 50,000 RK4 steps; 200,000 demagnetization calls & 2.55 & -- & 0.570 & 128 / 7.34 \\
\bottomrule
\end{tabular}}
\end{table*}

\begin{figure*}[!tp]
\centering
\includegraphics[width=\textwidth]{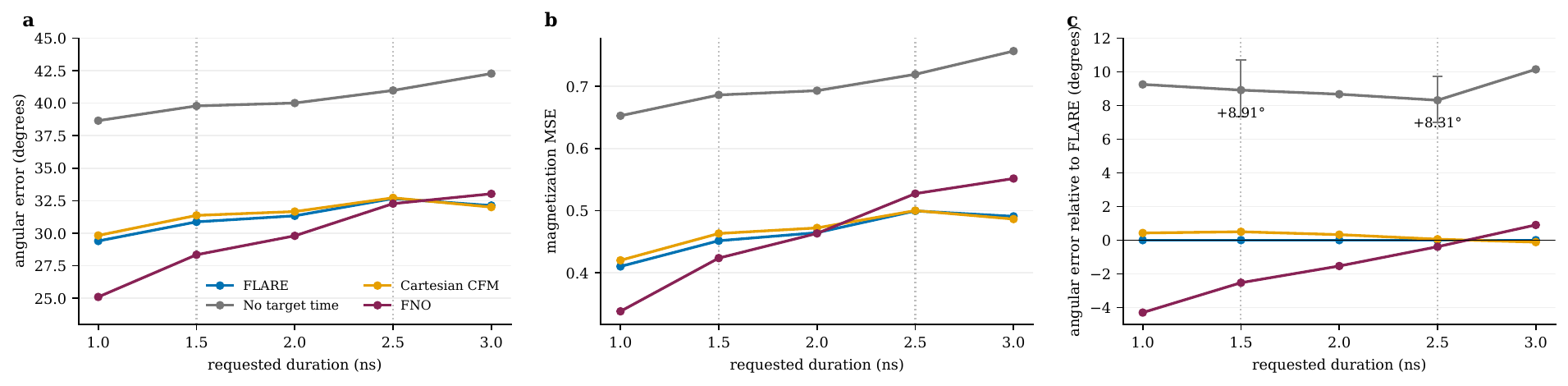}
\caption{\textbf{Conditioning on the requested physical horizon.}
Dotted lines mark unseen 1.5- and 2.5-ns interpolation queries.
\textbf{(a)} Angular error.  \textbf{(b)} magnetization MSE.
\textbf{(c)} angle relative to Stage~1 on the same cases; whiskers at the
two unseen horizons are paired 95\% base-condition bootstrap intervals.
Removing the
explicit target-time channels consistently worsens the matched prediction,
including at both unseen horizons.}
\label{fig:app_time_interpolation}
\end{figure*}

Primary Stage~1 ran for 7:35:57 on two H200s, or 15.20 device-hours.
The GT-only, generated-state-only, and mixed Stage~2 continuations consumed
11.30, 12.06, and 11.52 additional H200-hours, respectively; the Stage~1 plus
three-continuation matrix therefore used 50.07 H200-hours. The six external
endpoint-model 50,000 runs total 26.89 H200-hours, and their common 10,000 bring-up
runs add 5.42 H200-hours. NeuralMAG's reported 50,000 component adaptation used
2.55 H200-hours; its excluded short pilots used another 0.36 H200-hours.
These totals exclude MuMax$^3$ dataset generation, which is shared by every
learned method, and separate representation/OOD/ensemble training used only for
ablations.
Using the measured $B=1$ times, the primary Stage~1 training cost is amortized
after
\[
\frac{15.198\ {\rm H200\,h}\times3600}
     {446.125\ {\rm s}-0.995204\ {\rm s}}
=122.9
\]
standardized fixed two-segment 5-ns endpoint queries, i.e., approximately 123 queries. Counting
Stage~1 plus one mixed Stage~2 continuation raises this training-only
break-even to approximately 216 queries. These are compute amortization
figures, not claims that complete-path quality was evaluated at a uniform
5-ns physical duration.
Figure~\ref{fig:app_time_interpolation} shows the matched duration sweep and
the effect of masking target-time conditioning.

At 1.5 and 2.5\,ns, explicit target-time conditioning improves the matched no-time control by $8.91^\circ$ [7.32, 10.69] and $8.31^\circ$ [6.97, 9.73], respectively. Across all 1,493 single-segment test transitions, masking target time increases angular error by $7.625^\circ$ [6.64, 8.71]. These are interpolation queries within the realized constant-control duration support. The post-drive segment is nominally 3.5\,ns, but its boundary and the 0.25-ns saved-frame grid leave no exact 3.5-ns held-out pair; exact targets are available through 3.0\,ns on the tested grid. This saved-frame restriction does not apply to the predicted handoff in Table~\ref{tab:main_distribution}. Requested 4--6\,ns one-shot rows cross a control boundary because the configured 1--6\,ns interval is an absolute-time filter. We report these unsupported rows explicitly rather than filling them by extrapolation.
\subsection{Complete supported-horizon distribution audit}
\label{app:horizon_distribution}

\begin{figure*}[!tp]
\centering
\includegraphics[width=\textwidth]{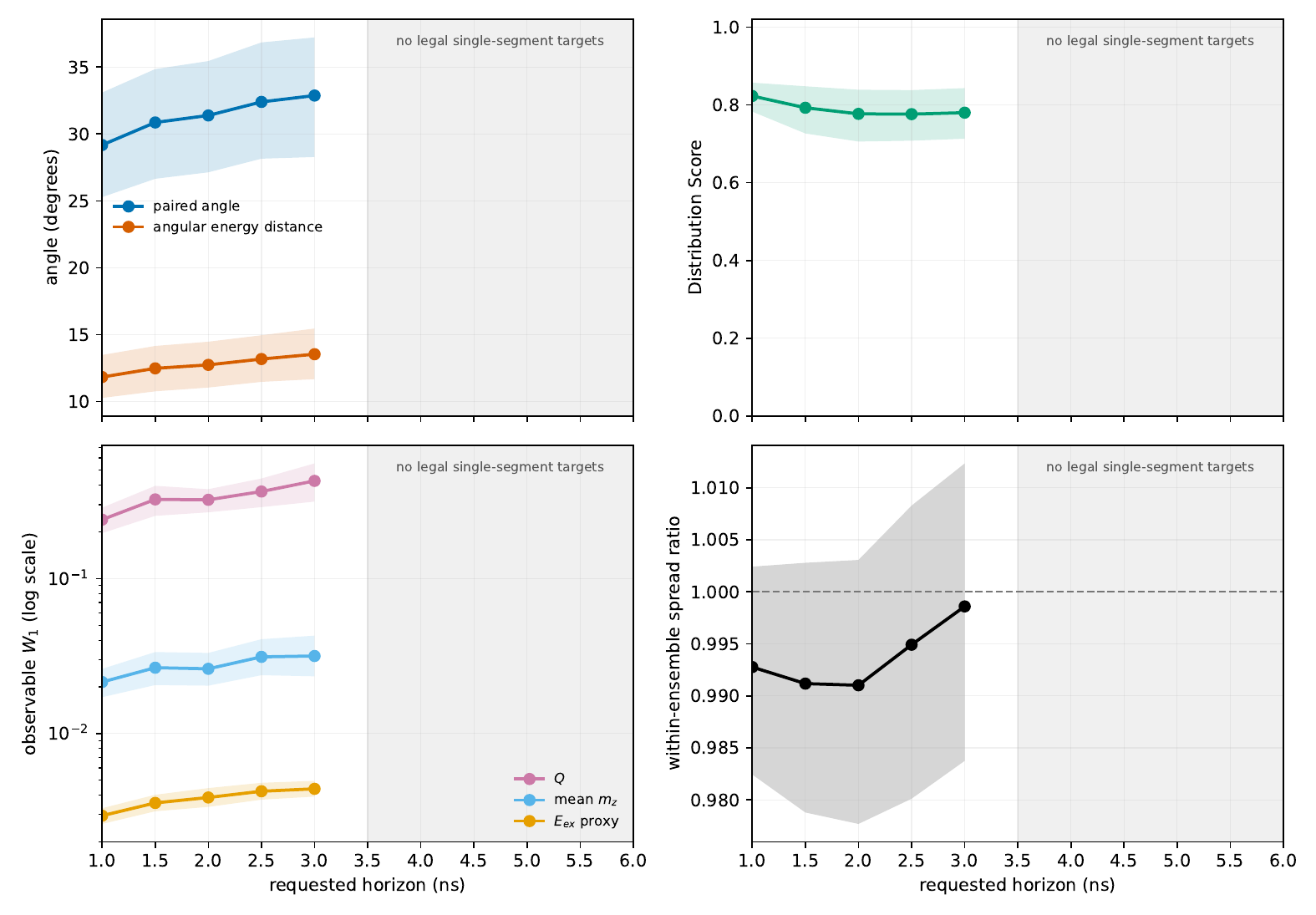}
\caption{\textbf{Complete supported-horizon diagnostics.} Circles at 1, 2,
and 3~ns lie on the training duration grid; 1.5 and 2.5~ns are held-out
interpolation queries. \textbf{(a)} Repeat-paired angular error and angular
energy distance. \textbf{(b)} Distribution Score. \textbf{(c)} Empirical
Wasserstein-1 error for $Q$, mean $m_z$, and the normalized exchange-texture
energy proxy. \textbf{(d)} Ratio of model to MuMax$^3$ within-ensemble angular
spread; the dashed line is one. Bands are 95\% base-cluster intervals. The
gray region is shown explicitly because it contains no legal exact
single-segment target in the held-out saved-frame data.}
\label{fig:app_horizon_diagnostics}
\end{figure*}

We evaluate the frozen 50,000 checkpoint at every exact target available on the
requested integer and half-integer grid. Each supported row contains 33
held-out base groups from the post-drive relaxation branch; the driven pulse
is shorter than 1~ns and therefore supplies no row on this grid. For every
condition, 25 forecasts comprise five stochastic draws from each of the five
exact MuMax$^3$ anchors. Distribution Score, angular energy distance, and the
physical-observable distributions use an equal five-versus-five subset (one
forecast per anchor), while fair energy score uses all five draws separately
for each exact anchor. Intervals resample base groups and keep all anchors and
draws together.
Figure~\ref{fig:app_horizon_diagnostics} summarizes all supported horizons
and explicitly marks unsupported requests.
Table~\ref{tab:app_horizon_distribution} gives the distribution metrics and
support status at every requested horizon.
\begin{table*}[!tp]
\centering
\scriptsize
\setlength{\tabcolsep}{2.8pt}
\caption{Distribution metrics at every requested integer and half-integer
horizon from 1 to 6~ns. Brackets are 95\% base-cluster intervals; arrows
give the favorable direction. ``Seen'' means that the duration lies on the
training duration grid, not that the held-out physical condition was seen.
Unsupported rows are retained rather than extrapolated.}
\label{tab:app_horizon_distribution}
\resizebox{\textwidth}{!}{%
\begin{tabular}{lclcccc}
\toprule
Horizon (ns) & Support & $N_{\rm base}$ & Distribution Score $\uparrow$ & Paired angle ($^\circ$) $\downarrow$ & Angular energy dist. ($^\circ$) $\downarrow$ & Fair energy score $\downarrow$ \\
\midrule
1.0 & seen & 33 & 0.823 [0.783, 0.857] & 29.17 [25.28, 33.09] & 11.83 [10.27, 13.48] & 0.3189 [0.2820, 0.3559] \\
1.5 & interpolation & 33 & 0.793 [0.727, 0.848] & 30.86 [26.65, 34.85] & 12.48 [10.76, 14.15] & 0.3397 [0.3008, 0.3780] \\
2.0 & seen & 33 & 0.777 [0.706, 0.839] & 31.38 [27.13, 35.44] & 12.74 [11.05, 14.47] & 0.3446 [0.3058, 0.3822] \\
2.5 & interpolation & 33 & 0.776 [0.708, 0.838] & 32.39 [28.15, 36.84] & 13.18 [11.47, 14.95] & 0.3594 [0.3207, 0.3986] \\
3.0 & seen & 33 & 0.780 [0.713, 0.843] & 32.87 [28.27, 37.21] & 13.54 [11.68, 15.46] & 0.3622 [0.3225, 0.4020] \\
\midrule
3.5 & no exact saved target & 0 & -- & -- & -- & -- \\
4.0 & outside one segment & 0 & -- & -- & -- & -- \\
4.5 & outside one segment & 0 & -- & -- & -- & -- \\
5.0 & outside one segment & 0 & -- & -- & -- & -- \\
5.5 & outside one segment & 0 & -- & -- & -- & -- \\
6.0 & outside one segment & 0 & -- & -- & -- & -- \\
\bottomrule
\end{tabular}}
\end{table*}

Table~\ref{tab:app_horizon_physics_spread} gives the corresponding physical
observables and within-ensemble spread.
\begin{table*}[!tp]
\centering
\scriptsize
\setlength{\tabcolsep}{3.0pt}
\caption{Physical-observable distribution errors and ensemble spread over
the five supported horizons. Observable columns are empirical
five-versus-five Wasserstein-1 errors. Model and MuMax$^3$ spread are the
equal-condition means of within-set angular distances in degrees. Brackets
are 95\% base-cluster intervals.}
\label{tab:app_horizon_physics_spread}
\resizebox{\textwidth}{!}{%
\begin{tabular}{lcccccc}
\toprule
Horizon (ns) & $W_1(Q)$ & $W_1(\overline m_z)$ & $W_1(E_{\rm ex}\ \mathrm{proxy})$ & Model spread ($^\circ$) & MuMax$^3$ spread ($^\circ$) & Spread ratio [95\% CI] \\
\midrule
1.0 & 0.24 [0.20, 0.29] & 0.0215 [0.0171, 0.0261] & 0.00294 [0.00261, 0.00329] & 49.28 & 49.64 & 0.993 [0.982, 1.002] \\
1.5 & 0.32 [0.25, 0.40] & 0.0266 [0.0205, 0.0335] & 0.00356 [0.00315, 0.00402] & 48.85 & 49.28 & 0.991 [0.979, 1.003] \\
2.0 & 0.32 [0.27, 0.38] & 0.0261 [0.0204, 0.0331] & 0.00386 [0.00336, 0.00444] & 48.46 & 48.90 & 0.991 [0.978, 1.003] \\
2.5 & 0.37 [0.29, 0.44] & 0.0312 [0.0238, 0.0406] & 0.00424 [0.00374, 0.00480] & 48.05 & 48.30 & 0.995 [0.980, 1.008] \\
3.0 & 0.43 [0.31, 0.55] & 0.0316 [0.0234, 0.0428] & 0.00439 [0.00391, 0.00494] & 47.51 & 47.58 & 0.999 [0.984, 1.012] \\
\bottomrule
\end{tabular}}
\end{table*}

Across the legal 1--3~ns range, the paired angle increases smoothly from
$29.17^\circ$ to $32.87^\circ$, angular energy distance from $11.83^\circ$
to $13.54^\circ$, and fair energy score from 0.3189 to 0.3622; Distribution
Score remains between 0.776 and 0.823. The two interpolation rows lie on the
same gradual trend rather than showing an off-grid discontinuity. Physical
observable errors also increase with horizon, most clearly for $Q$ and the
exchange-texture proxy. This is controlled degradation, not horizon-invariant
accuracy.

The model/MuMax$^3$ spread ratio ranges only from 0.991 to 0.999, and every 95\%
interval contains one. Thus, over the evaluated 1--3~ns support, we find no
statistically resolved mode collapse or over-diffusion. This conclusion is
restricted to these post-relaxation rows: the empty 3.5--6~ns entries cannot
support any claim about one-shot long-horizon fidelity.
\section{Multi-seed representation and objective controls}
\label{app:seed_geometry}
The controls in Table~\ref{tab:design_ablation}(a) share the split, architecture, global batch 128, 50,000 updates, and seeds 78--80. The core uses local axis--angle fields; Cartesian CFM transports normalized spins;
the 2D variant uses a fixed tangent basis; Riemannian FM follows a geodesic bridge \citep{chen2024flow}; and Direct U-Net directly regresses $\Omega_\star$ with MSE.

\paragraph{Additional complete-path controls.} Table~\ref{tab:design_ablation}(b) re-evaluates FLARE, Cartesian CFM, and no-time at seed 78 with frozen checkpoint conditioning and fixed cuDNN kernel selection. Saved fields are normalized before scoring. Five repeat-matched endpoint errors are averaged within each of 33 bases; gains use 50,000 paired base-bootstrap resamples, conditional on the checkpoints. Persistence retains the drive-start field. The original dimensionless energy diagnostic uses unit exchange and zero DMI, anisotropy, and encoded field; it is not the physical MuMax$^3$ total energy.
Five-seed ensembles combine seeds 78--82 of Direct U-Net and four external baselines, with 50,000 updates per member and no test-set selection. Quality uses each member once across the five repeat anchors per condition, assigned independently of targets; fair energy uses all five members at each exact anchor. FLARE uses five draws from one seed-78 checkpoint. Each forecast retains its own predicted handoff.
\begin{table*}[!tp]
\centering
\small
\caption{Five-seed ensembles (78--82) on the exact-control complete path, with 95\% base-cluster intervals conditional on those members; the frozen FLARE seed-78 reference is repeated for comparison. Each path applies the driven control from its saved drive-start state, uses the predicted boundary for the exact 3.5-ns zero-current relaxation, and scores the final endpoint. Distribution Score, angular energy distance, and paired angle use 33 base conditions with matched five-versus-five endpoints; fair energy uses 165 exact anchors. The three-seed representation/objective controls are reported in Table~\ref{tab:design_ablation}(a).}
\label{tab:app_seed_geometry}
\resizebox{\textwidth}{!}{%
\begin{tabular}{lcccc}
\toprule
Variant & Distribution Score $\uparrow$ & Angular energy dist. ($^\circ$) $\downarrow$ & Paired angle ($^\circ$) $\downarrow$ & Fair energy score $\downarrow$ \\
\midrule
FLARE (one checkpoint) & \textbf{0.762} [0.660, 0.844] & \textbf{15.30} [13.32, 17.26] & 36.72 [31.93, 41.38] & \textbf{0.393} [0.356, 0.430] \\
Direct U-Net & 0.472 [0.188, 0.778] & 20.91 [17.89, 24.24] & 38.88 [34.61, 43.04] & 0.541 [0.499, 0.580] \\
CNO-FM & 0.717 [0.619, 0.799] & 20.52 [17.23, 24.15] & 35.14 [31.28, 38.94] & 0.535 [0.488, 0.582] \\
DPOT-Ti & 0.744 [0.660, 0.818] & 25.86 [20.06, 32.43] & 40.76 [35.23, 46.34] & 0.569 [0.510, 0.628] \\
Poseidon-T & 0.703 [0.610, 0.780] & 18.93 [16.06, 22.04] & \textbf{33.31} [29.12, 37.41] & 0.503 [0.456, 0.548] \\
PDEArena U-Net & 0.515 [0.264, 0.760] & 24.86 [19.55, 30.91] & 38.96 [33.38, 44.56] & 0.558 [0.493, 0.623] \\
\bottomrule
\end{tabular}%
}
\end{table*}

\begin{figure*}[!tp]
\centering
\includegraphics[width=\textwidth]{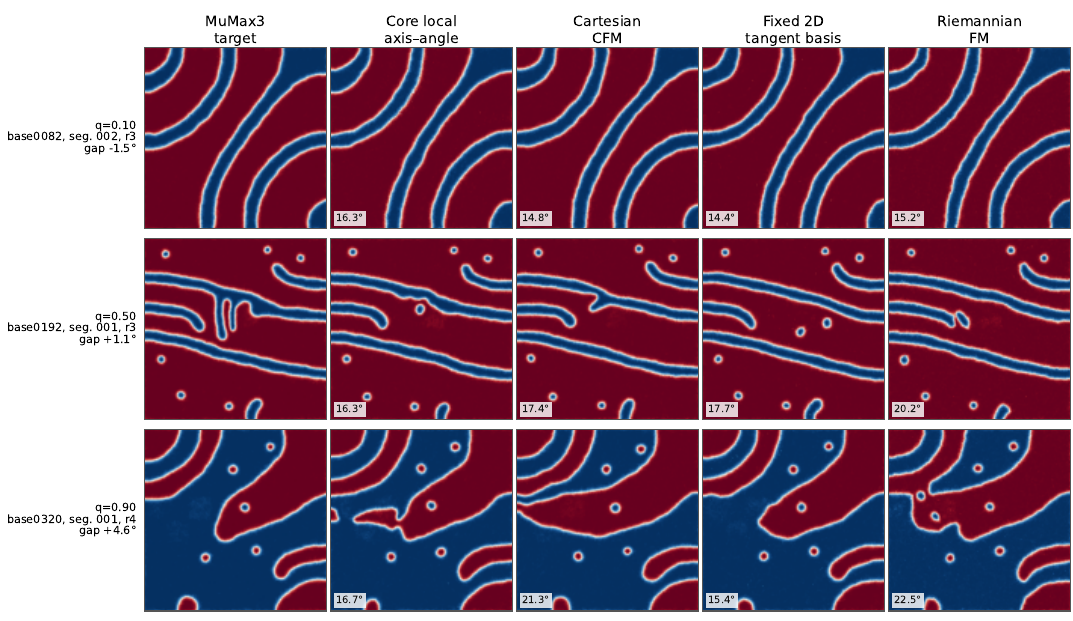}
\caption{\textbf{Separate single-segment qualitative geometry-ablation predictions.}
Rows are the 10th, 50th, and 90th percentiles of the seed-78
Cartesian-CFM-minus-core exact-anchor mean-draw angular-error gap over 330
anchors, selected without visual inspection. Each method panel shows the
medoid of five draws under inter-draw angular distance; inset values are mean
draw-to-MuMax$^3$ angular errors. Cartesian CFM is competitive in the low-gap
case but shows larger domain-wall displacement in the upper tail. All panels
show $m_z\in[-1,1]$.}
\label{fig:app_seed_geometry_predictions}
\end{figure*}

Figure~\ref{fig:app_seed_geometry_predictions} gives a separate
single-segment, quantile-selected view of the seed-78 prediction fields; it is
not a visualization of the complete-path rollout rows in
Table~\ref{tab:design_ablation}(a).
The core varies little across seeds. Its mean is better than Cartesian CFM, Riemannian FM, and Direct U-Net on all four metrics. The first two support the rotation coordinates; Direct U-Net shows that the generative objective matters beyond the shared coordinates and backbone. %
The fixed 2D basis is close in mean but has substantially greater seed variation. Together these results support the coordinate family while separating that conclusion from prior- or basis-specific claims.
\paragraph{Five-model ensembles.}
At a matched five-output budget, one FLARE model achieves 19.2\% lower angular energy distance ($15.30^\circ$ versus $18.93^\circ$) and 21.8\% lower fair energy (0.393 versus 0.503) than the best external five-model ensemble, Poseidon-T. Its Distribution Score is 0.762 versus 0.744 for the best external ensemble on that metric, DPOT-Ti (+2.4\%). These compare one FLARE fit with five independently trained fits per ensemble, not equal total training compute. Poseidon-T's ensemble has a lower paired angle ($33.31^\circ$ versus $36.72^\circ$), preserving the distinction between realization-level and distributional accuracy.

Ensembling improves fair energy substantially, but Direct U-Net's five-model ensemble still scores 0.541 versus FLARE's 0.393 despite sharing its backbone and rotation target. We also evaluate all ten three-member subsets of each method's five checkpoints, without selecting a subset for the official comparison. All 50 subsets remain worse than FLARE on fair energy; the lowest subset score is 0.490. These subset results test sensitivity to member choice, whereas Table~\ref{tab:app_seed_geometry} retains the matched five-output comparison. Its bootstrap intervals condition on the fixed members and measure variation across held-out base conditions, not uncertainty from choosing a new set of training seeds.
\paragraph{Anchor-jitter sensitivity.}
For the same five deterministic models at training seed 78, we perturb the
initial drive-start anchor with independent Gaussian fields smoothed at
$\sigma=4$ pixels. Noise is projected into each spin's tangent plane, rescaled
to a prescribed pre-decoding valid-cell RMS tangent-space rotation magnitude, and decoded by the $S^2$ exponential map;
nonmagnetic cells remain zero. We sweep 18 amplitudes from $0^\circ$ to
$80^\circ$ without retraining. Only the initial anchor is perturbed: each
forecast then follows its own predicted handoff, with unchanged controls and
horizons. All scores use the original, unperturbed MuMax$^3$ reference endpoints,
so the perturbation is an internal randomization of the predictive method rather
than a change to the reference task. Fair energy uses five independently
perturbed forecasts per exact anchor for nonzero jitter and a one-point
distribution at zero jitter.

Table~\ref{tab:app_anchor_jitter} gives the complete sweep. The response is
strongly non-monotone: performance initially improves for every deterministic
baseline, before degrading at larger angles. Direct U-Net reaches
the best score in the sweep, 0.5022 at $40^\circ$; the minima of CNO-FM,
DPOT-Ti, Poseidon-T, and PDEArena U-Net occur at $55^\circ$, $55^\circ$,
$55^\circ$, and $50^\circ$, respectively. Nevertheless, even the best
perturbed deterministic result remains above FLARE's unperturbed fair energy of
0.393 reported in Table~\ref{tab:main_distribution}.

\begin{table*}[!tp]
\centering
\small
\setlength{\tabcolsep}{4pt}
\caption{\textbf{Fine-grained anchor-jitter sensitivity in fair energy.}
Each deterministic method uses its seed-78 checkpoint; jitter is the prescribed pre-decoding valid-cell
RMS tangent-space rotation magnitude applied only at the initial anchor. Values are point estimates. Lower is better. Bold denotes the best result for each method across all jitter amplitudes. FLARE is not perturbed and has fair
energy 0.393 (Table~\ref{tab:main_distribution}).}
\label{tab:app_anchor_jitter}
\resizebox{\textwidth}{!}{%
\begin{tabular}{rccccc}
\toprule
Jitter ($^\circ$) & Direct U-Net & CNO-FM & DPOT-Ti & Poseidon-T & PDEArena U-Net \\
\midrule
0    & 0.8077 & 0.7451 & 0.8140 & 0.7515 & 0.7743 \\
0.25 & 0.7914 & 0.7437 & 0.8123 & 0.7481 & 0.7721 \\
0.5  & 0.7819 & 0.7425 & 0.8109 & 0.7453 & 0.7710 \\
1    & 0.7633 & 0.7401 & 0.8083 & 0.7402 & 0.7675 \\
2    & 0.7338 & 0.7355 & 0.8035 & 0.7312 & 0.7638 \\
5    & 0.6778 & 0.7233 & 0.7873 & 0.7062 & 0.7502 \\
10   & 0.6242 & 0.7055 & 0.7645 & 0.6800 & 0.7263 \\
20   & 0.5622 & 0.6706 & 0.7295 & 0.6409 & 0.6851 \\
30   & 0.5219 & 0.6383 & 0.6977 & 0.6032 & 0.6489 \\
35   & 0.5075 & 0.6206 & 0.6775 & 0.5916 & 0.6244 \\
40   & \textbf{0.5022} & 0.5927 & 0.6357 & 0.5764 & 0.5797 \\
45   & 0.5055 & 0.5670 & 0.5941 & 0.5652 & 0.5511 \\
50   & 0.5235 & 0.5455 & 0.5707 & 0.5487 & \textbf{0.5381} \\
55   & 0.5460 & \textbf{0.5246} & \textbf{0.5661} & \textbf{0.5398} & 0.5493 \\
60   & 0.5684 & 0.5299 & 0.5939 & 0.5422 & 0.5736 \\
65   & 0.5927 & 0.5473 & 0.6549 & 0.5558 & 0.5941 \\
70   & 0.6137 & 0.5768 & 0.7277 & 0.5679 & 0.6183 \\
80   & 0.6453 & 0.6251 & 0.8463 & 0.5885 & 0.6479 \\
\bottomrule
\end{tabular}%
}
\end{table*}

\section{Multi-segment composition}
\label{app:composition}
\begin{figure*}[!tp]
\centering
\includegraphics[width=\textwidth]{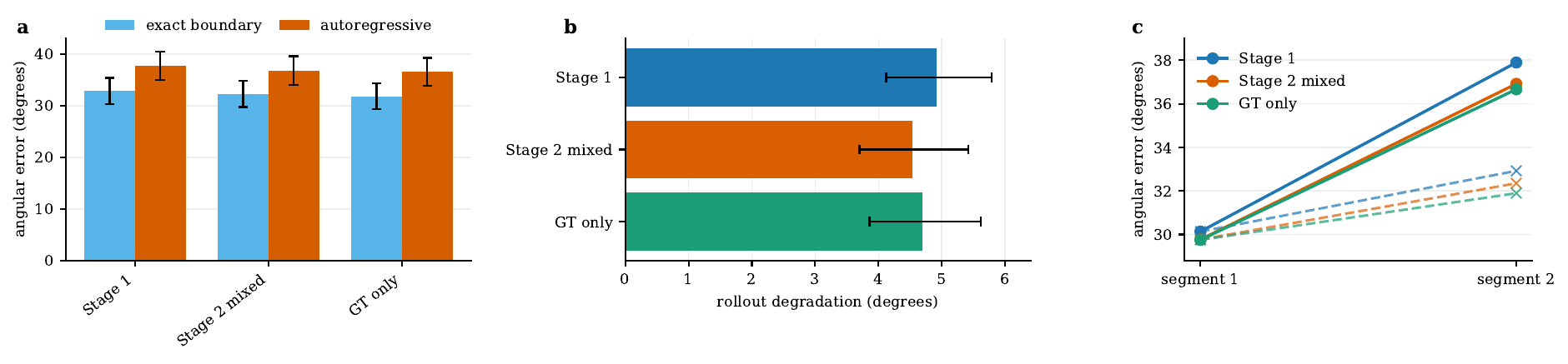}
\caption{\textbf{Teacher-forced and autoregressive rollouts.}
\textbf{(a)} Angular error for the compute-matched Stage--2 controls.
\textbf{(b)} rollout degradation, defined as autoregressive minus teacher-forced
error. \textbf{(c)} segment-resolved error for Stage~1, mixed Stage~2, and
GT-only Stage~2. The first segment shares the real initial state; subsequent
rollout steps expose each method to model-generated states. Intervals cluster 736
rollouts by 97 base conditions.}
\label{fig:app_rollout_diagnostics}
\end{figure*}

\begin{figure*}[!tp]
    \centering
\includegraphics[width=\linewidth]{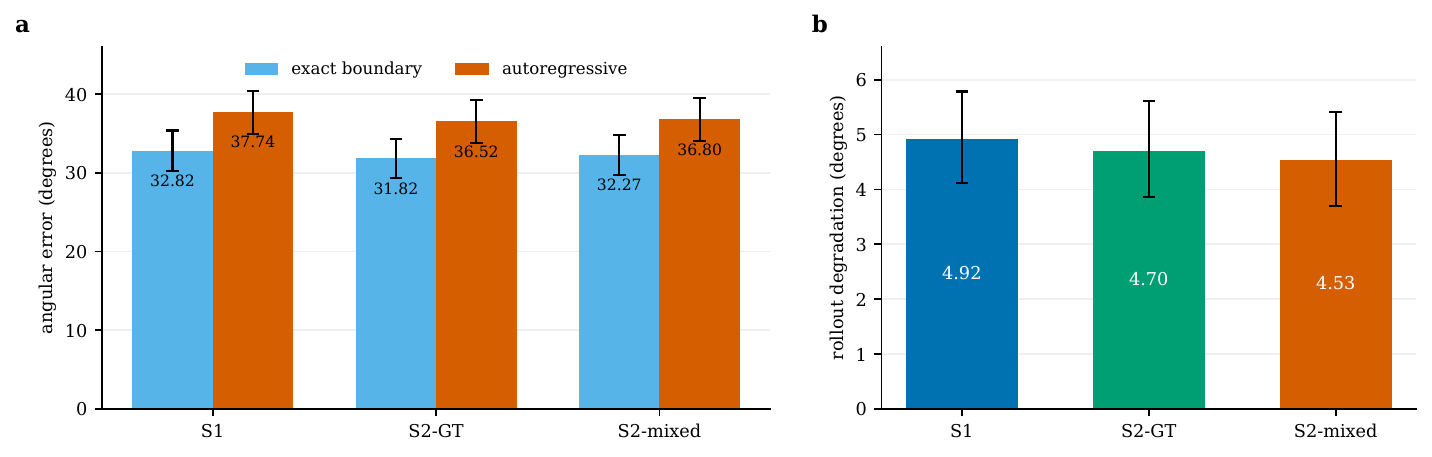}
    \caption{Stage-2 multi-stage evaluation using the same formal results as
    Table~\ref{tab:app_stage2}.  \textbf{(a)} Exact-boundary and autoregressive angular
    error.  \textbf{(b)} Autoregressive minus exact-boundary error.
    Whiskers are 95\% base-condition bootstrap intervals.}
    \label{fig:stage2_results}
\end{figure*}

Figure~\ref{fig:app_rollout_diagnostics} gives the aggregate rollout
diagnostics, and Figure~\ref{fig:stage2_results} provides the compact Stage--2
comparison.

Table~\ref{tab:app_stage2} reports the compute-matched numerical results.
\begin{table*}[!tp]\centering
\small
\caption{Multi-segment evaluation and compute-matched Stage--2 controls. Brackets are 95\% base-condition cluster-bootstrap intervals; lower is better.}
\label{tab:app_stage2}
\resizebox{\textwidth}{!}{%
\begin{tabular}{lccc}
\toprule
Method & Ground-truth-boundary angle ($^\circ$) & Rollout angle ($^\circ$) & Rollout degradation ($^\circ$) \\
\midrule
FLARE-S1 & 32.82 [30.27, 35.38] & 37.74 [34.95, 40.44] & 4.92 [4.12, 5.79] \\
FLARE-S2-GT & \textbf{31.82} [29.30, 34.33] & \textbf{36.52} [33.80, 39.22] & 4.70 [3.86, 5.62] \\
FLARE-S2-mixed & 32.27 [29.73, 34.81] & 36.80 [34.05, 39.55] & \textbf{4.53} [3.70, 5.42] \\
\bottomrule
\end{tabular}%
}
\vspace{3pt}
\resizebox{\textwidth}{!}{%
\begin{tabular}{lcc}
\toprule
Paired comparison & Rollout error reduction (\%) $\uparrow$ & Rollout degradation reduction (\%) $\uparrow$ \\
\midrule
S2-mixed vs. S1 & 2.48 [1.49, 3.45] & 7.87 [0.80, 14.57] \\
S2-mixed vs. S2-GT & $-$0.78 [$-$1.51, $-$0.09] & 3.61 [$-$0.94, 7.99] \\
\bottomrule
\end{tabular}%
}
\end{table*}

\begin{figure*}[!tp]
\centering
\includegraphics[width=\textwidth]{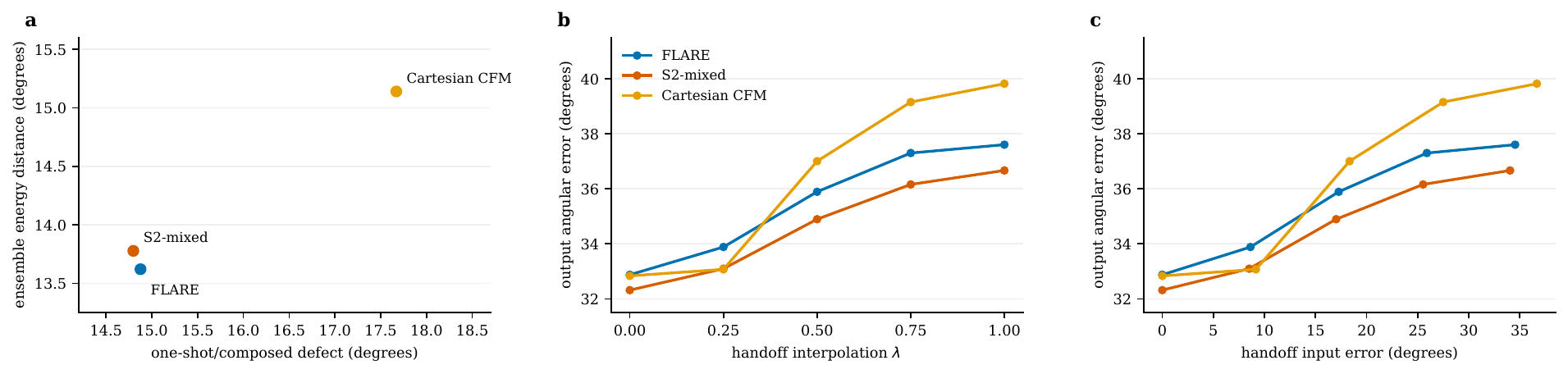}
\caption{\textbf{Composition consistency and boundary sensitivity.}
\textbf{(a)} One-shot/composed defect versus the energy distance between the
two output ensembles.  \textbf{(b)} final error as the second-segment input
moves from the exact boundary ($\lambda=0$) to a generated boundary
($\lambda=1$).  \textbf{(c)} the same curves parameterized by measured input
error.  Internal consistency does not imply physical fidelity: a small path
defect can coexist with a worse ensemble distance.}
\label{fig:app_semigroup_handoff}
\end{figure*}

Relative to Stage~1, S2-mixed reduces autoregressive rollout error by $2.48\%$ [1.49\%, 3.45\%] and rollout degradation by $7.87\%$ [0.80\%, 14.57\%]. Relative to equal-budget S2-GT, however, S2-mixed has $0.78\%$ [0.09\%, 1.51\%] higher autoregressive error, while its 3.61\% degradation reduction has a confidence interval crossing zero [$-0.94\%$, 7.99\%]. %
Mixed rollout-state exposure is therefore not superior to GT-only continuation on absolute rollout error, and we do not treat it as a separate contribution.
Figure~\ref{fig:app_semigroup_handoff} visualizes composition consistency and
the continuous boundary-perturbation response.

For two adjacent spans with identical non-duration controls, the semigroup
test compares one one-shot prediction over the summed horizon with two
composed samples. FLARE Stage~1 has a $14.88^\circ$
[13.58, 16.17] mean one-shot/composed defect and a $13.62^\circ$
[12.51, 14.70] ensemble energy distance. The rotation model reduces these
two quantities by $2.79^\circ$ [2.35, 3.24] and $1.52^\circ$ [1.20, 1.86]
relative to matched Cartesian CFM. This internal-consistency diagnostic
cannot replace comparison with MuMax$^3$.

Boundary sensitivity follows the unit-sphere interpolation
\begin{equation}
\begin{gathered}
m_\lambda=\mathcal R\!\left(
\lambda\,\operatorname{Log}_{m_{\rm exact}}(m_{\rm pred})
\right)m_{\rm exact},
\\ \lambda\in\{0,0.25,0.5,0.75,1\}.
\end{gathered}
\label{eq:app_handoff}
\end{equation}
For mixed Stage~2, final error grows smoothly from $32.32^\circ$ at the exact
boundary to $36.67^\circ$ at the predicted boundary, with no abrupt failure
on the sampled path.

\section{Leave-one-mask-family-out: held-out ring controls}
\label{app:ood}
\begin{figure*}[!tp]
\centering
\includegraphics[width=\textwidth]{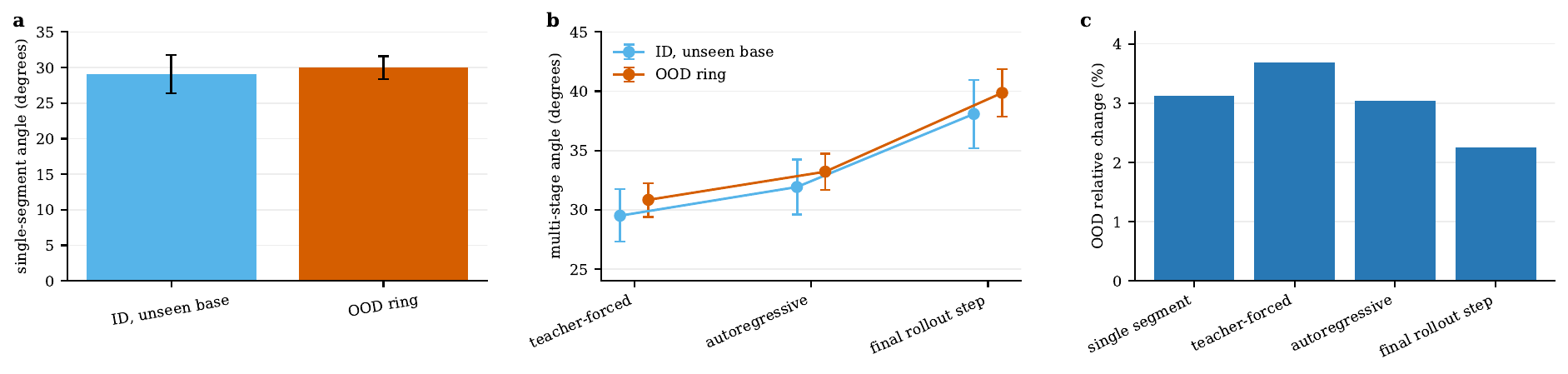}
\caption{\textbf{Generalization to a geometry family excluded before
training.} \textbf{(a)} FLARE ID and ring-OOD angular errors.
\textbf{(b)} the matched auxiliary FNO architecture control (not a
Table~\ref{tab:main_distribution} comparator).  \textbf{(c)} relative OOD changes for
single-segment, teacher-forced, autoregressive, and final-rollout-step
evaluation.  Error bars resample base conditions separately within each
stratum.}
\label{fig:app_ood_ring}
\end{figure*}

Figure~\ref{fig:app_ood_ring} summarizes the held-out-ring shift and matched
auxiliary FNO control.
Table~\ref{tab:ring_ood} compares ID and ring-OOD performance on the non-ring ID and held-out ring OOD sets. 
\begin{table*}[!tp]
\centering
\caption{Generalization to the completely held-out ring geometry. Brackets are 95\% base-condition cluster-bootstrap intervals.}
\label{tab:ring_ood}

\begin{tabular}{lcccc}
\toprule
Evaluation & ID angle ($^\circ$)$\downarrow$ & Ring-OOD angle ($^\circ$)$\downarrow$ & $\Delta$ ($^\circ$) & $\Delta$ (\%) \\
\midrule
Single segment
& 29.05 [26.42, 31.73]
& 29.96 [28.36, 31.59]
& +0.91
& +3.12 \\

Autoregressive handoff
& 31.40 [28.64, 34.15]
& 32.35 [30.64, 34.05]
& +0.95
& +3.04 \\
\bottomrule
\end{tabular}
\end{table*}

Table~\ref{tab:app_ood_ring_numeric} gives the corresponding numerical
intervals for every evaluation mode.
\begin{table*}[!tp]
\centering
\scriptsize
\setlength{\tabcolsep}{3.2pt}
\caption{FLARE under the ring-family leave-one-factor-out split. The ID
column uses unseen non-ring bases from the same refitted split; the OOD column
contains only ring controls, all excluded before fitting and normalization.
Intervals resample base conditions.}
\label{tab:app_ood_ring_numeric}
\begin{tabular}{lcc}
\toprule
Evaluation & Non-ring ID [95\% CI] & Held-out ring [95\% CI] \\
\midrule
Single-segment angle ($^\circ$) & 29.05 [26.42, 31.73] & 29.96 [28.36, 31.59] \\
Single-segment MSE & 0.42 [0.37, 0.48] & 0.44 [0.41, 0.47] \\
Single-segment $|\Delta Q|$ & 0.54 [0.477, 0.60] & 0.56 [0.52, 0.60] \\
Teacher-forced rollout angle ($^\circ$) & 28.85 [26.14, 31.60] & 29.92 [28.26, 31.59] \\
Autoregressive rollout angle ($^\circ$) & 31.40 [28.64, 34.15] & 32.35 [30.64, 34.05] \\
Final-step angle ($^\circ$) & 35.42 [32.28, 38.60] & 36.21 [34.28, 38.13] \\
\bottomrule
\end{tabular}
\end{table*}

This is a true leave-one-factor-out experiment rather than a random test
subset: all ring base conditions are removed before fitting the model and
normalization statistics. The remaining split contains 633 training and 79
validation base groups; evaluation uses 79 non-ring ID groups and 209 ring
groups.
Complete autoregressive rollouts are available for 74 ID and 198 OOD
base groups. FLARE's ring shift is $+0.91^\circ$ (3.12\%) in single-segment evaluation,
$+0.95^\circ$ (3.04\%) over complete autoregressive rollouts, and
$+0.80^\circ$ (2.25\%) at the final segment. ID and OOD confidence intervals
overlap substantially. We therefore describe the result as stable transfer
under one entirely unseen mask family, not proof of invariant performance.

The experiment keeps resolution, simulator, representation, and broad
physical-parameter support fixed.  It does not establish zero-shot transfer
to three-dimensional samples, arbitrary meshes, unseen boundary laws, or
physical parameters far outside training support.

\section{Zero-shot generalization to unseen temperatures}
\label{app:temperature_ood}

We evaluate the frozen Stage~1 FLARE checkpoint at two temperatures that are absent from training and validation but lie within the training temperature range. The checkpoint is
trained at 30, 150, and 300~K, while 90~K and 225~K are the midpoints of the
adjacent 30--150~K and 150--300~K intervals, respectively. The model weights,
EMA state, training normalization statistics, sampler, and evaluation protocol
are kept fixed throughout the experiment.

For each unseen temperature, we use the same 36 held-out test base conditions
and generate a fresh 1-ns MuMax relaxation followed by five independent thermal
continuations per base condition. Single-segment and distributional metrics use
all 36 matched test bases; autoregressive handoff uses 35 because one base lacks
an eligible two-segment trajectory. Distributional metrics compare five FLARE
samples with five MuMax repeats under each matched physical condition.

Prediction difficulty varies systematically with temperature. We therefore use,
for each matched base condition, the arithmetic midpoint of the corresponding
scalar metrics at the two adjacent training temperatures as the ID interpolation
reference. We compare each unseen temperature with this reference using 10,000
paired base-condition cluster-bootstrap replicates. For angular error and angular
energy distance, degradation is defined as unseen minus reference; for
Distribution Score, degradation is defined as reference minus unseen. Positive
values therefore indicate worse performance at the unseen temperature.

\begin{table*}[!tp]
\centering
\small
\setlength{\tabcolsep}{3.5pt}
\caption{\textbf{Zero-shot generalization to in-range temperatures absent during
training.}
The frozen Stage~1 checkpoint is trained at 30, 150, and 300~K. For 90~K and
225~K, the ID reference is the paired per-base midpoint of the two adjacent
training temperatures. Brackets report 95\% paired base-condition cluster-bootstrap
intervals. Positive degradation denotes worse performance at the unseen
temperature.}
\label{tab:temperature_ood_inrange}
\resizebox{\textwidth}{!}{%
\begin{tabular}{clrrrr}
\toprule
$T$ & Metric & ID interpolation reference & Unseen temperature
& Absolute degradation [95\% CI] & Relative degradation [95\% CI] \\
\midrule
90 K
& Single-segment angle ($^\circ$)
& 23.815 & 24.118
& $+0.303$ [$-0.230$, $+0.843$]
& $+1.27\%$ [$-0.96\%$, $+3.55\%$] \\

90 K
& Autoregressive handoff angle ($^\circ$)
& 31.213 & 29.917
& $-1.296$ [$-2.573$, $-0.117$]
& $-4.15\%$ [$-8.06\%$, $-0.39\%$] \\

90 K
& Distribution Score
& 0.6687 & 0.7380
& $-0.0693$ [$-0.1616$, $+0.0270$]
& $-10.36\%$ [$-24.99\%$, $+3.99\%$] \\

90 K
& Angular energy distance ($^\circ$)
& 12.749 & 12.385
& $-0.365$ [$-1.388$, $+0.598$]
& $-2.86\%$ [$-10.50\%$, $+4.89\%$] \\

\midrule

225 K
& Single-segment angle ($^\circ$)
& 37.291 & 37.903
& $+0.612$ [$-0.102$, $+1.300$]
& $+1.64\%$ [$-0.28\%$, $+3.45\%$] \\

225 K
& Autoregressive handoff angle ($^\circ$)
& 44.320 & 45.372
& $+1.052$ [$-0.781$, $+2.777$]
& $+2.37\%$ [$-1.77\%$, $+6.36\%$] \\

225 K
& Distribution Score
& 0.9123 & 0.9151
& $-0.0028$ [$-0.0238$, $+0.0188$]
& $-0.30\%$ [$-2.62\%$, $+2.06\%$] \\

225 K
& Angular energy distance ($^\circ$)
& 16.754 & 16.926
& $+0.172$ [$-0.748$, $+1.123$]
& $+1.03\%$ [$-4.45\%$, $+6.78\%$] \\
\bottomrule
\end{tabular}%
}
\end{table*}

At 225~K, all four metrics remain statistically consistent with the adjacent
150--300~K interpolation reference. At 90~K, single-segment error, Distribution
Score, and angular energy distance are likewise consistent with the 30--150~K
reference, while autoregressive handoff improves by 4.15\% [0.39\%, 8.06\%].
Across both unseen in-range temperatures, the frozen FLARE model therefore
follows the temperature-dependent ID trend with no temperature-specific
adaptation.

A more demanding zero-shot test at 400~K probes extrapolation beyond the
maximum training temperature of 300~K. Relative to the matched 300~K endpoint,
single-segment and autoregressive angular errors increase by 17.54\%
[15.00\%, 20.18\%] and 16.75\% [12.58\%, 21.21\%], respectively, while angular
energy distance increases by 44.24\% [35.47\%, 53.20\%]. Together, the in-range
and out-of-range tests show strong interpolation across unseen temperatures
inside the training support and a clear loss of fidelity under temperature
extrapolation beyond that support.

\section{Magnetization-field audit}
\label{app:visual}

\subsection{Selection and display conventions}

Single-segment examples are the cases nearest the 25th, 50th, and 90th
percentiles of Stage~1 angular error after averaging repeated model draws.
Autoregressive examples use the median and 90th-percentile final-hop Stage~1
errors.  The semigroup and boundary-interpolation plates use the numerically
median Stage~1 defect and mixed-Stage~2 boundary error, respectively.
Selection is performed before any image is rendered.

\begin{figure*}[!tp]
\centering
\includegraphics[width=0.96\textwidth]{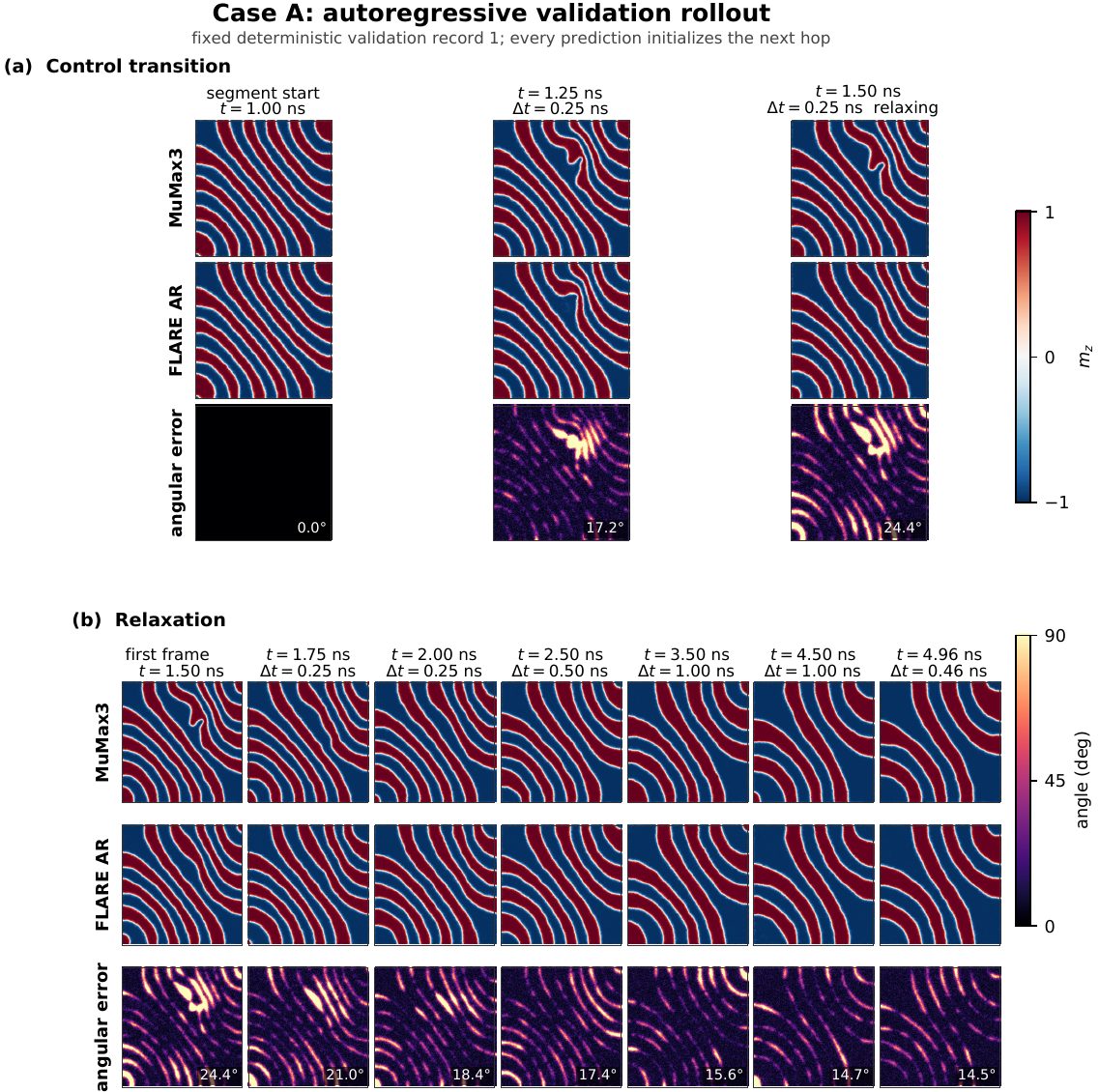}
\caption{\textbf{Time-resolved autoregressive rollout A: difficult stripe and ring-wall reorganization.} MuMax$^3$, the FLARE rollout, and the full-vector angular error are aligned at every displayed time.  The first relaxation frame has $24.4^\circ$ mean error; the longer relaxation reaches $14.5^\circ$.  Error localizes on moving walls and newly resolved branches, while the large-scale magnetic regime remains recognizable.}
\label{fig:app_magnetic_timeseries_a}
\end{figure*}

\begin{figure*}[!tp]
\centering
\includegraphics[width=0.96\textwidth]{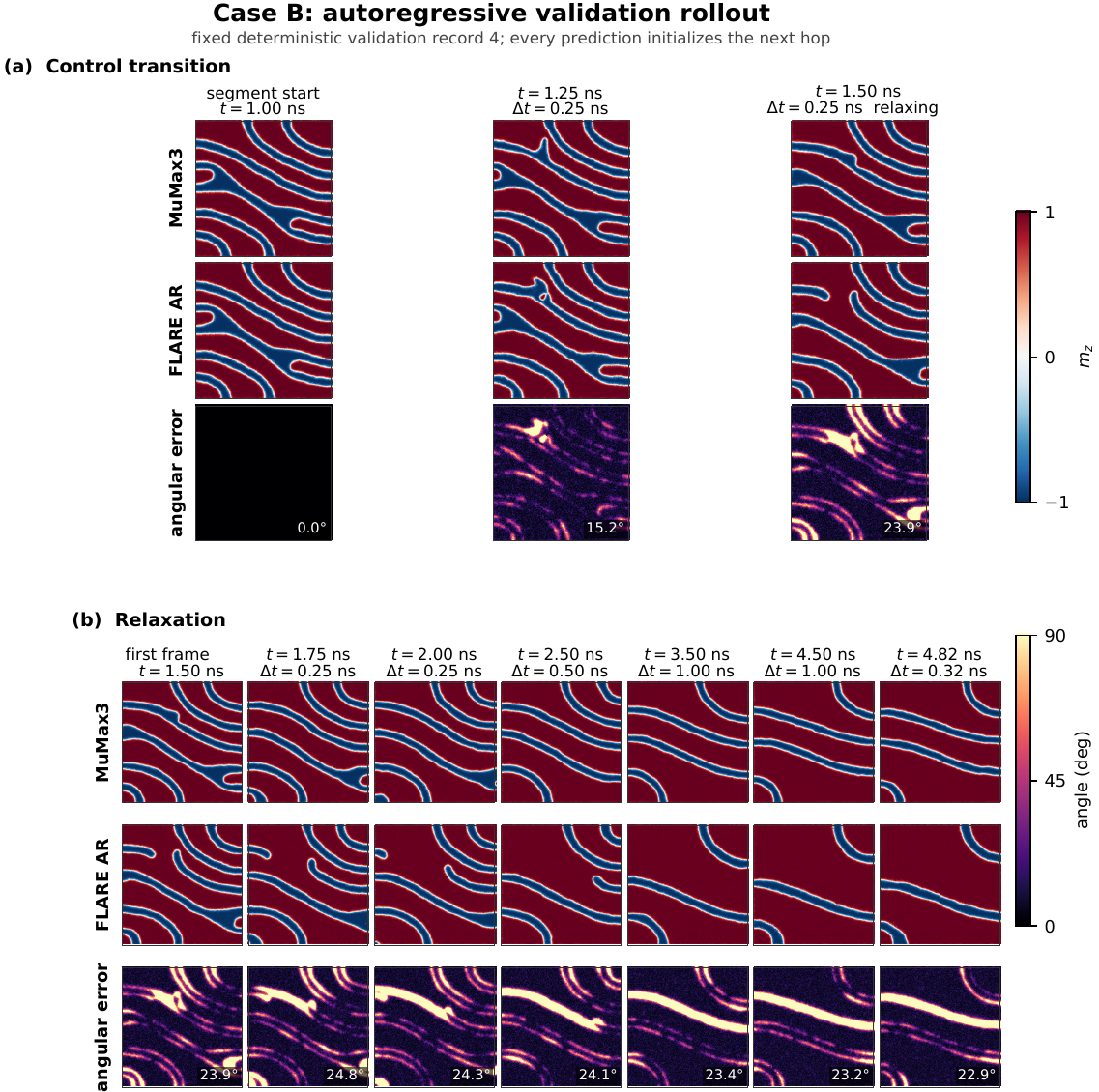}
\caption{\textbf{Time-resolved autoregressive rollout B: stable stripe-wall evolution.}
The same fully autoregressive protocol is applied to a second fixed case.
Wall motion and the disappearance of small domains remain qualitatively similar over
the long second segment; its final mean angle is $22.9^\circ$, compared with
$23.9^\circ$ in the first relaxation frame.}
\label{fig:app_magnetic_timeseries_b}
\end{figure*}

\begin{figure*}[!tp]
\centering
\includegraphics[width=0.96\textwidth]{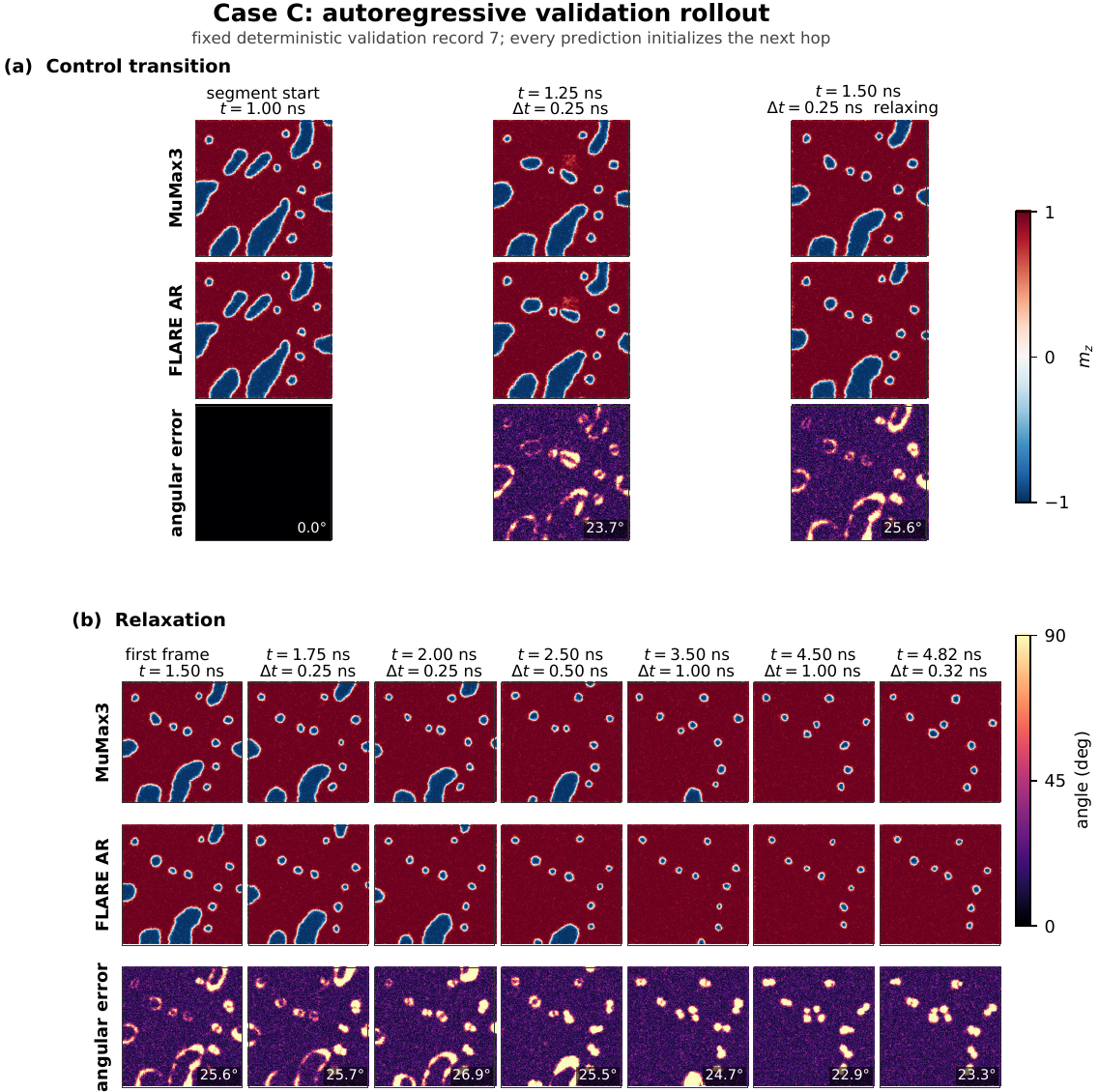}
\caption{\textbf{Time-resolved autoregressive rollout C: many-body domain extinction.}
The model follows the removal and contraction of several small reversed
domains across both controls.  Its errors remain concentrated at interfaces
and surviving cores instead of spreading through saturated regions; the
first-relaxation and final means are $25.6^\circ$ and $23.3^\circ$.}
\label{fig:app_magnetic_timeseries_c}
\end{figure*}

\begin{figure*}[!tp]
\centering
\includegraphics[width=0.96\textwidth]{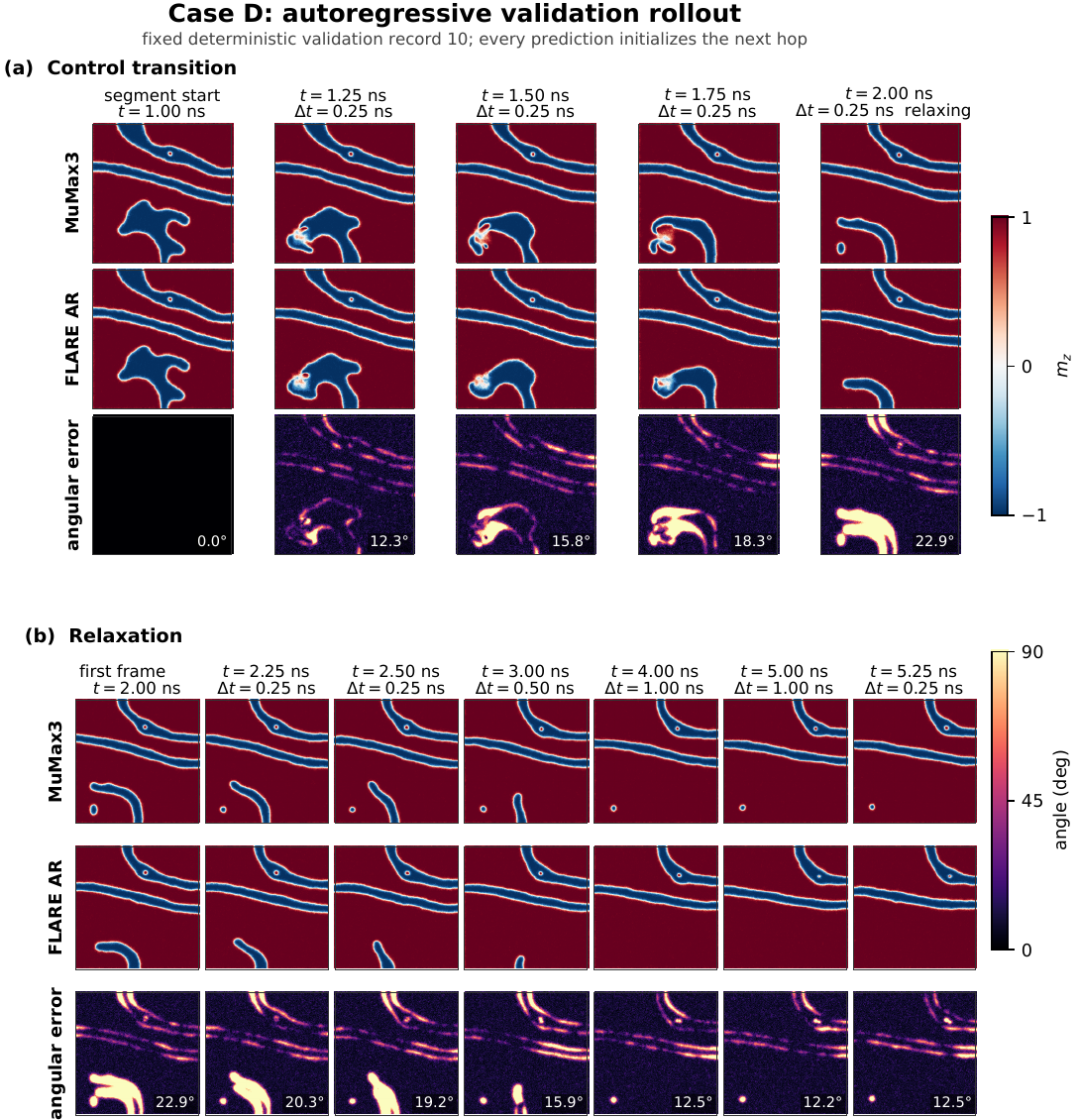}
\caption{\textbf{Time-resolved autoregressive rollout D: large-domain contraction.}
FLARE preserves the two long walls while a lower domain retracts and leaves
the field of view.  The full-vector map reveals thin registration errors
that are hard to see from $m_z$ alone; mean angular error stays near
$12$--$23^\circ$ throughout the rollout.}
\label{fig:app_magnetic_timeseries_d}
\end{figure*}

Figures~\ref{fig:qualitative_single} and~\ref{fig:qualitative_ar}--\ref{fig:app_magnetic_handoff}
use the frozen formal case manifest and fixed visualization seeds.  Their
printed errors describe the displayed draw, whereas the selection quantiles
average repeated evaluation draws.  The median single-segment case is the
same as in Figure~\ref{fig:app_magnetic_baselines}.
Figures~\ref{fig:app_magnetic_timeseries_a},
\ref{fig:app_magnetic_timeseries_b},
\ref{fig:app_magnetic_timeseries_c}, and~\ref{fig:app_magnetic_timeseries_d}
use deterministic validation record indices 1, 4, 7, and 10 with the seed-78 Stage~1 checkpoint (Heun-10); they are fixed independently of error and were not screened visually. Each plate follows one case through both controls, with every prediction used to initialize the next hop. Queries crossing a control change are split at the exact protocol switch, including fractional substeps; displayed times are saved-frame times, not control boundaries. These diagnostics add within-segment queries to the two-query main-table path.

All magnetization panels use the same fixed $m_z\in[-1,1]$ scale.  Angular
maps show the full three-component spin angle and clip only the display at
$90^\circ$; printed means use the unclipped values.  Black arrows subsample
the in-plane components.  The topology panel visualizes the finite-difference
residual of $m\cdot(\partial_xm\times\partial_ym)/(4\pi)$ and is diagnostic
rather than a separate evaluation metric.

\subsection{Time-resolved autoregressive rollouts}

\begin{figure*}[!tp]
\centering
\includegraphics[width=\textwidth]{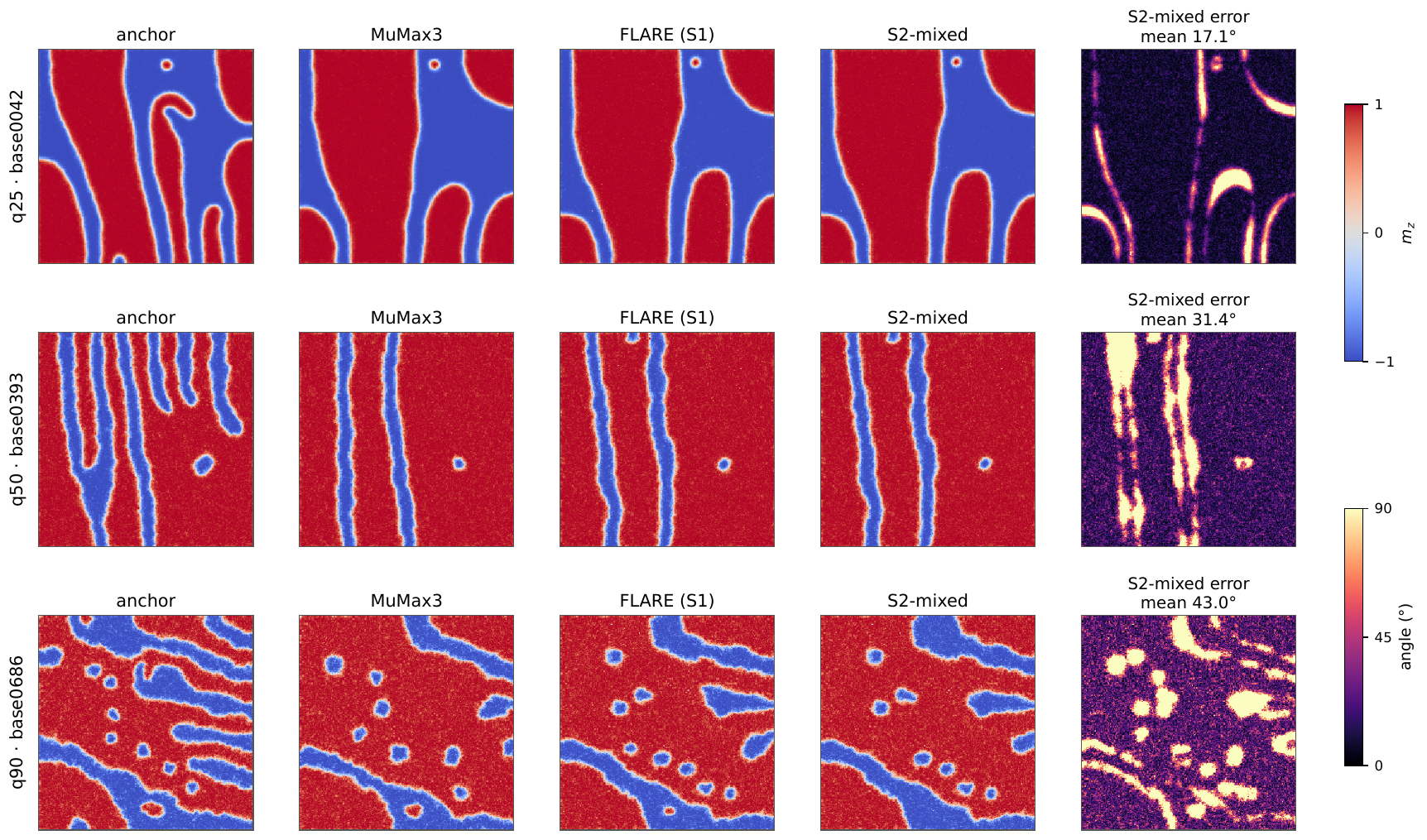}
\caption{\textbf{Single-segment magnetic evolution across numerical
difficulty quantiles.} Rows are selected at the 25th percentile, median, and
90th percentile without visual inspection.  Columns show the anchor,
MuMax$^3$ target, Stage~1, S2-mixed, and the Stage~2
three-component angular-error map.  Stage~2 mean angles for these displayed
draws are $17.1^\circ$, $31.4^\circ$, and $43.0^\circ$.  Wall displacement
and changes in small domains become more pronounced in the difficult case.}
\label{fig:qualitative_single}
\end{figure*}

Figure~\ref{fig:qualitative_single} compares fixed single-segment difficulty
quantiles.

\begin{figure*}[!tp]
\centering
\includegraphics[width=\textwidth]{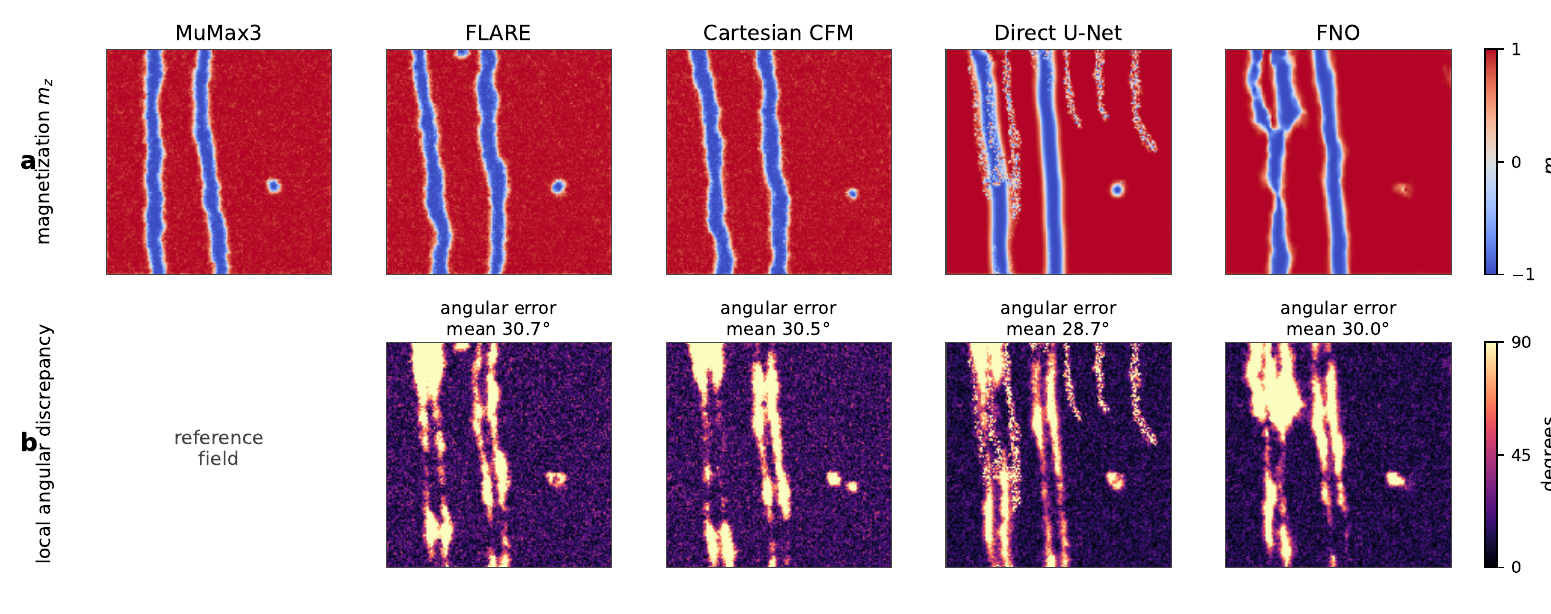}
\caption{\textbf{Method-level field and error comparison on the fixed median
case.} The top row shows $m_z$; the bottom row shows per-cell full-vector
angular error relative to the same MuMax$^3$ target.  The methods retain
different amounts of fine-scale wall roughness.
The auxiliary internal FNO control's localized defects are more visible in
the angle map than in the scalar state average; FNO is not a
Table~\ref{tab:main_distribution} comparator.  This plate is illustrative;
aggregate rankings use all held-out cases.}
\label{fig:app_magnetic_baselines}
\end{figure*}

Figure~\ref{fig:app_magnetic_baselines} compares representative method-level fields and local errors on the formal seed-78 FLARE median case.

\begin{figure*}[!tp]
\centering
\includegraphics[width=0.93\textwidth]{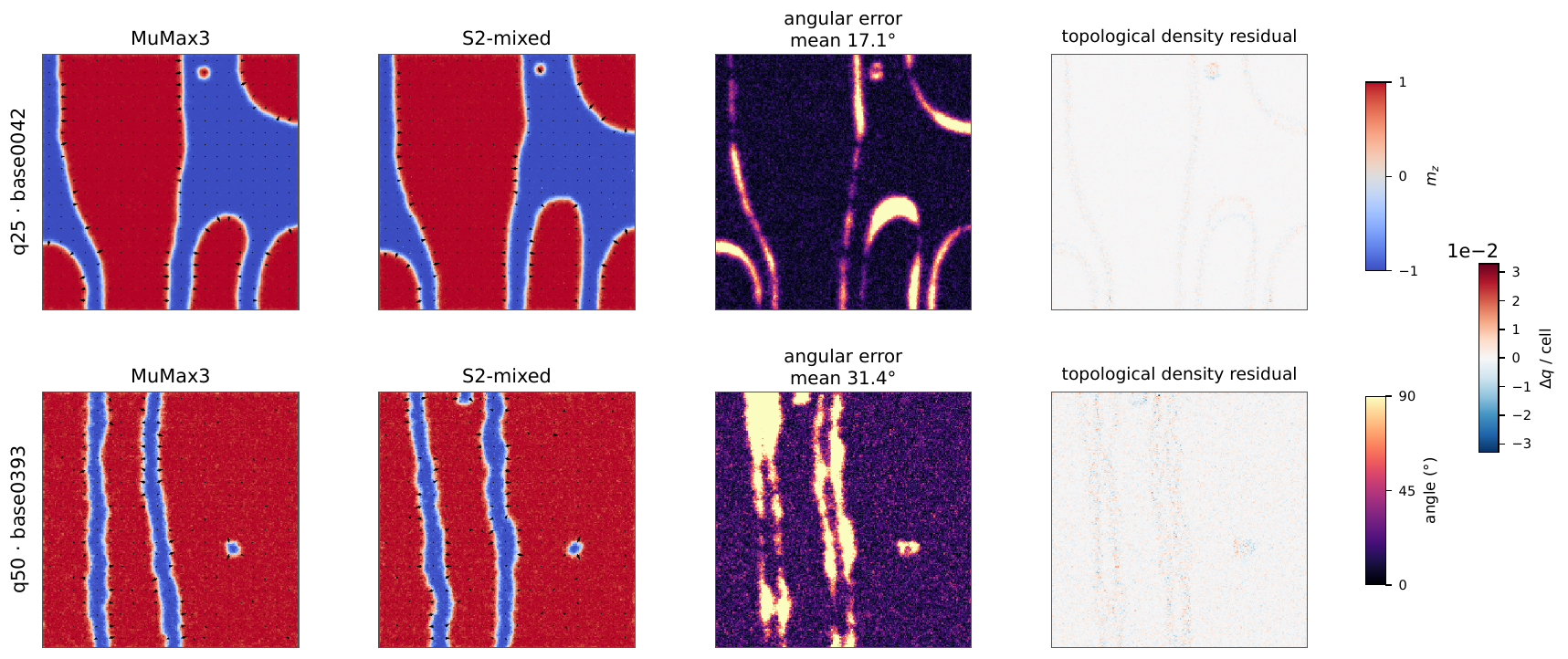}
\caption{\textbf{Vector texture and local topology.} The formal q25 and median
cases from Figure~\ref{fig:qualitative_single} are reused without further
selection.  MuMax$^3$ and S2-mixed are shown with sparse in-plane arrows over
$m_z$.  The third column gives full-vector angular error; the fourth gives
predicted-minus-reference topological density, computed by open-boundary
central differences in grid-cell units on a shared symmetric color scale.
The angle maps highlight displaced walls and changes in small domains;
local topological residuals retain both positive and negative contributions.}
\label{fig:app_magnetic_topology}
\end{figure*}

Figure~\ref{fig:app_magnetic_topology} relates vector texture errors to local
topological-density residuals.

\begin{figure*}[!tp]
\centering
\includegraphics[width=\textwidth]{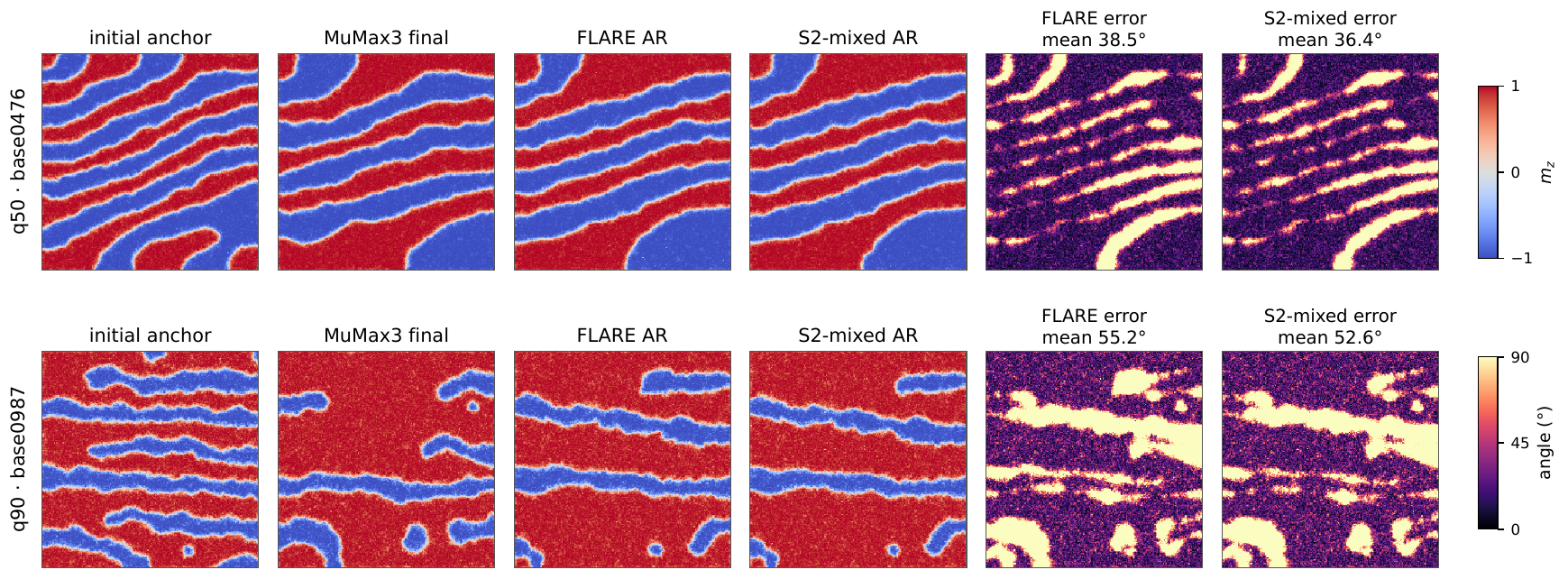}

\caption{\textbf{Fully autoregressive magnetic-field rollouts.} Rows are the
median and 90th-percentile rollouts under the fixed numerical selection
rule.  Stage~1 and mixed Stage~2 receive their own first-segment prediction
as the next anchor.  For these displayed draws, the mean final error decreases
from $38.5^\circ$ to $36.4^\circ$ at the median and from $55.2^\circ$ to
$52.6^\circ$ in the hard tail.  These two examples do not establish a uniform
improvement over all cases or over the equal-budget GT-only control.
The final two columns show full-vector angular errors relative to MuMax$^3$.}
\label{fig:qualitative_ar}
\end{figure*}

Figure~\ref{fig:qualitative_ar} shows the selected fully autoregressive
rollouts.

\begin{figure*}[!tp]
\centering
\includegraphics[width=\textwidth]{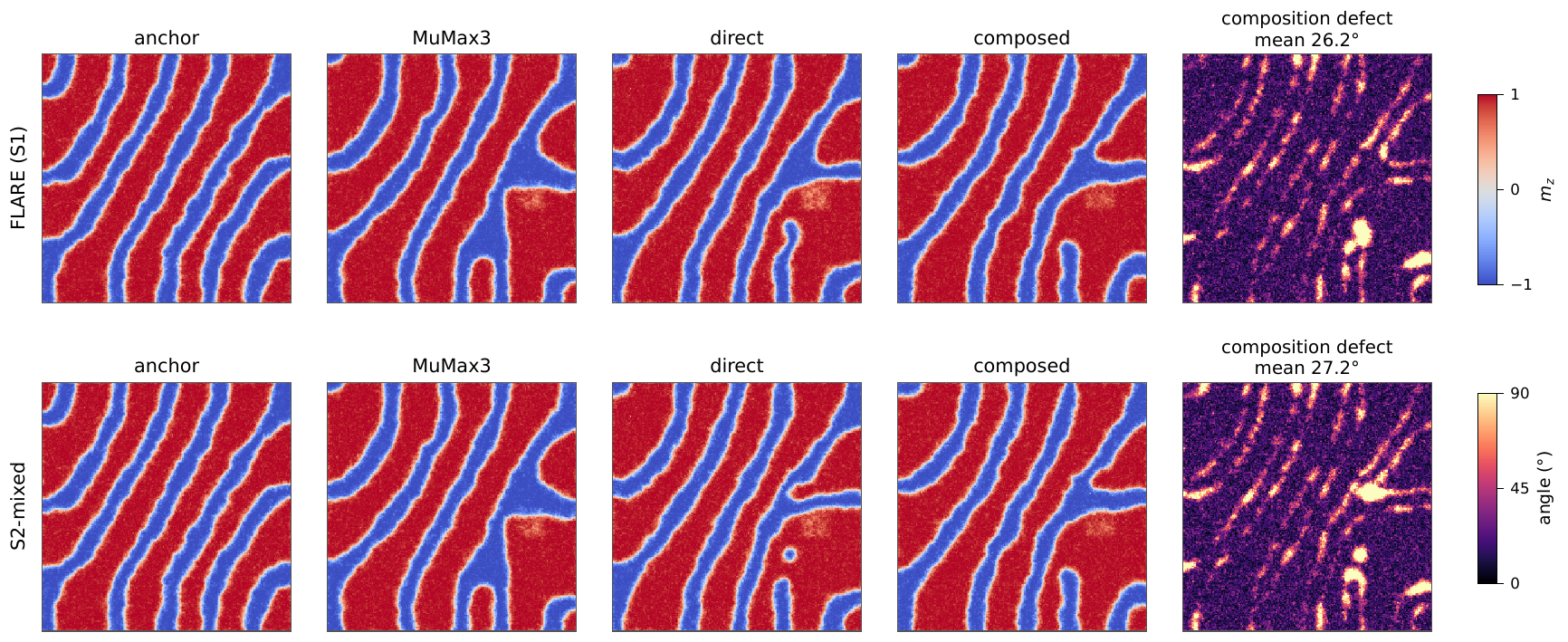}
\caption{\textbf{One-shot versus composed magnetic transitions.} For the
fixed median-defect case, each row compares a direct transition over the
summed horizon with two adjacent generated transitions.  The rightmost map
is their local composition defect.  The selected case has a $15.15^\circ$
mean Stage~1 defect over the evaluation draws; the separate displayed draws
have defects of $26.2^\circ$ (Stage~1) and $27.2^\circ$ (S2-mixed).
Both routes retain a stripe pattern but disagree on wall placement and
branch structure.  Agreement between model routes does not establish
agreement with MuMax$^3$.}
\label{fig:qualitative_semigroup}
\end{figure*}

Figure~\ref{fig:qualitative_semigroup} compares one-shot and composed magnetic
transitions.

\begin{figure*}[!tp]
\centering
\includegraphics[width=\textwidth]{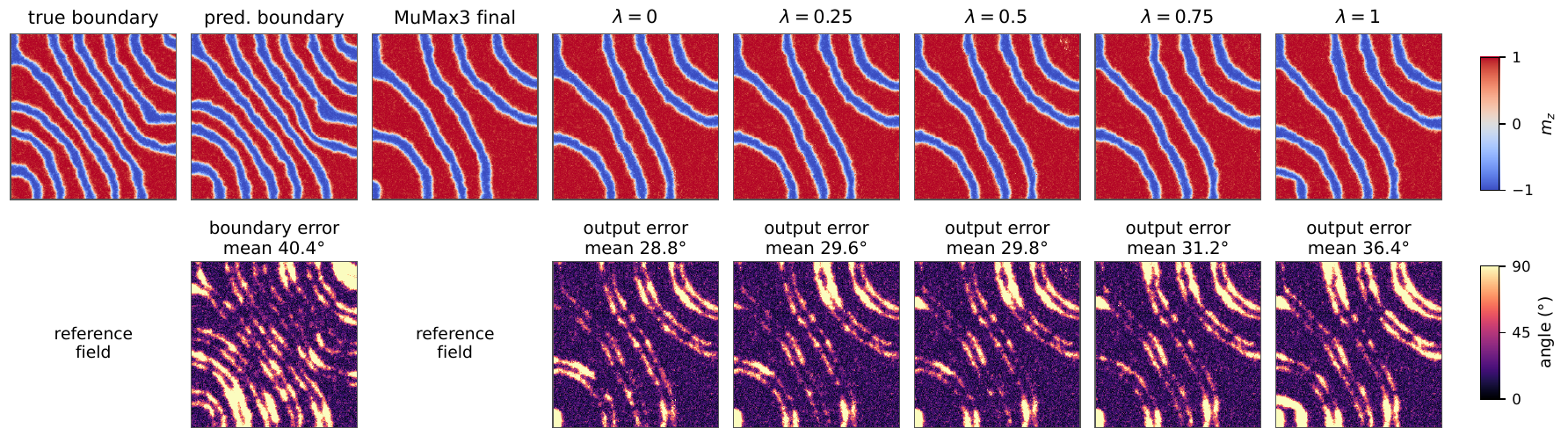}

\caption{\textbf{Continuous rollout-state perturbation in magnetization space.}
The first row shows the true and generated segment boundary, the final
MuMax$^3$ state, and Stage~2 predictions as $\lambda$ moves from the exact to the
predicted boundary.  The second row shows boundary or continuation-state angular
error.  The continuation sampling seed is held fixed across $\lambda$;
the displayed mean output error increases from $28.8^\circ$ at the true
boundary to $36.4^\circ$ at the predicted boundary.  The angle maps expose
changes around displaced walls along this unit-sphere interpolation path.}
\label{fig:app_magnetic_handoff}
\end{figure*}

Figure~\ref{fig:app_magnetic_handoff} visualizes the continuous rollout-state
perturbation path.

\clearpage
\section{Failure modes, negative results, and reproducibility}
\label{app:limits}

\paragraph{Where errors occur.}
The field plates show three recurring mechanisms: displacement of a
high-gradient wall, uncertainty in a small skyrmion or branch, and
amplification of an already imperfect segment boundary.  Pixelwise MSE can
penalize a translated but structurally plausible object strongly, which is
why the evaluation pairs field metrics with local-translation and ensemble
diagnostics.  Conversely, translation tolerance can hide a topology error,
which is why charge and local error maps are retained.

\paragraph{Negative and mixed results.}
More ODE steps improve charge error slightly but worsen angular error and
cost.  Rollout-state-only continuation reduces the exposure gap further than GT-only but
worsens clean-boundary accuracy; the 1:1 mixture is a compromise.  We report
these results rather than only the metrics that favor FLARE.
\paragraph{Physical scope.}
FLARE produces conditional time-horizon predictions and multi-stage autoregressive rollouts.
It does not replace a numerical trajectory when dense transient dynamics,
first-passage time, conservation-law accounting, or a new physical law is
required.  Ring OOD changes one geometry family, not dimensionality or
discretization. Distributional resolution is limited by five MuMax$^3$ repeats
per condition.
Primary cluster-bootstrap intervals condition on one trained checkpoint; Table~\ref{tab:design_ablation}(a) separately reports three-seed variation for the core model and internal controls.
\paragraph{Compute disclosure.}
Table~\ref{tab:app_compute_accounting} reports device-hours, memory high-water
marks, temporal call counts, and training-cost amortization. Evaluation is
single-device except where independent array tasks are scheduled. The
additional magnetic figures require only inference from frozen checkpoints;
no visual case is used for checkpoint selection.

\section{Positioning: scope of the comparison}
\label{app:positioning}

Every metric in this paper is computed on a full magnetization field, so a
method is scoreable here only if it returns one at a requested time. This is
the inclusion rule behind Table~\ref{tab:positioning-full}: classes marked
\emph{n/a} are excluded by output type rather than by selection, and every
class omitted from the empirical comparison is instead marked \emph{not compared}.

The classes are, in order: classical micromagnetic solvers
\citep{donahue1999oommf,vansteenkiste2014design,bisotti2020fidimag,bruckner2023magnum,moreels2026mumaxplus,abert2025neuralmag},
which integrate the LLG equation directly and provide both our training data
and our timing reference; hybrid learned solvers
\citep{cai2024fast,nonaka2026magnex,chern2026machine}, which replace expensive field
computations with neural surrogates but retain stepwise integration;
reduced-order and latent dynamics
\citep{kovacs2019learning,exl2020learning,exl2021lowrank,schaffer2022prediction},
which evolve compressed states and map them back to full magnetization fields,
but use deterministic latent-space time advancement; observable-level neural ODEs \citep{chen2022forecasting,dolui2026node},
which advance a low-dimensional state and do not return the magnetization
field; static-field and energy-minimization networks
\citep{kovacs2022pinn,schaffer2023physics,schaffer2024constraintfree}, which
solve a field or equilibrium problem rather than a forward temporal one;
texture recognition, classification, and parameter inversion
\citep{iakovlev2018supervised,deviatov2019recurrent,wang2021learning,feng2024classification},
which are not forward simulators; general PDE surrogates
\citep{raonic2023cno,herde2024poseidon,hao2024dpot,mccabe2024multiple,gupta2022towards,wu2022learning,lippe2023pderefiner},
which predict full fields and generally advance on a fixed step grid; most
are deterministic, while PDE-Refiner uses stochastic iterative refinement;
and generative dynamics models developed for other physical
domains
\citep{lienen2024turbulence,price2025gencast,jing2024generative,nam2025flow},
which natively return distributions over full states or trajectories but use
domain-specific temporal interfaces: GenCast and LiFlow advance in fixed-horizon
steps, MDGEN generates fixed-stride trajectory windows, and TurbDiff samples
states without an initial condition. None has a released magnetization model.

\begin{table*}[!tp]
\caption{Positioning by method class. \emph{Full state}: returns the full
spatial state (for magnetization, the full field $\mathbf{m}$) rather than a
low-dimensional observable. \emph{Direct}: generates a state at a requested finite
horizon without integrating intermediate LLG steps. \emph{Partial}: the requested-horizon interface is absent or method-specific within the class (latent-space advancement, fixed-step rollout, or fixed-length trajectory windows) rather than a uniform one-shot interface. \emph{Stoch.}: natively returns a
distribution over outcomes rather than a point prediction. \emph{Magn.}:
developed for magnetization dynamics. \emph{n/a} marks classes that cannot be
scored by the field-based metrics used here. \emph{Not compared} means that no result under the common protocol is reported. Citations for each class are
given in the preceding paragraph.}
\label{tab:positioning-full}
\centering
\small
\begin{tabular}{@{}lcccc l@{}}
\toprule
Method class & Full state & Direct & Stoch. & Magn. & Status \\
\midrule
Classical solvers                  & $\checkmark$ & $\times$      & $\checkmark$ & $\checkmark$ & reference \\
Hybrid learned solvers             & $\checkmark$ & $\times$      & $\checkmark$ & $\checkmark$ & compared \\
Reduced-order dynamics             & $\checkmark$ & partial       & $\times$     & $\checkmark$ & not compared \\
Observable-level neural ODE        & $\times$     & $\checkmark$  & $\times$     & $\checkmark$ & n/a \\
Static-field / energy minimization & --  & --            & $\times$     & $\checkmark$ & n/a \\
Texture recognition / inversion    & --           & --            & --           & $\checkmark$ & n/a \\
General PDE surrogates             & $\checkmark$ & partial       & $\times$     & $\times$     & compared \\
Generative dynamics (other domains)& $\checkmark$ & partial       & $\checkmark$ & $\times$     & none released \\
\midrule
\textbf{FLARE}                     & $\checkmark$ & $\checkmark$  & $\checkmark$ & $\checkmark$ & this work \\
\bottomrule
\end{tabular}
\end{table*}

We searched the micromagnetics, computational-physics, and ML-for-science
literature through September~2,~2026 for learned models of magnetization dynamics that
return a full field at a requested horizon without stepwise LLG integration,
including forward citations from the simulation packages and hybrid solvers
listed above. We found no prior model that also treats the future state as a
conditional distribution, and we welcome pointers to work we have missed.
Except for thermal-field NeuralMAG-x5, the external comparators in Table~\ref{tab:main_distribution} are all
deterministic per checkpoint: their forecast distribution is degenerate at each exact anchor.
To rule out this explanation, we train five-seed ensembles for four external baselines and the matched Direct U-Net control (Table~\ref{tab:app_seed_geometry}) and add input jitter (Table~\ref{tab:app_anchor_jitter}); under both non-degenerate predictive distributions, FLARE retains its fair-energy advantage. Matched Cartesian CFM, Riemannian FM, and fixed 2D tangent-basis controls
share FLARE's architecture, conditioning, budget, and data but change transport
coordinates. Direct U-Net shares the rotation coordinates and backbone but
replaces flow matching with direct regression, isolating the objective
(Tables~\ref{tab:main_distribution} and~\ref{tab:design_ablation}). A deterministic FNO of comparable size serves as an
additional internal architecture control in Appendices~\ref{app:ood}
and~\ref{app:visual}.

Two scoping notes. NeuralMAG is adapted and reported in
Table~\ref{tab:main_distribution}; MagneX falls in the same class but is not compared, as
NeuralMAG already serves as a representative of this class: MagneX similarly learns the demagnetization-field operator while retaining stepwise deterministic LLG integration. PDE-Refiner
\citep{lippe2023pderefiner} improves long-horizon rollout through stochastic diffusion-inspired refinement and supports sample-based predictive uncertainty, but remains stepwise; we therefore group it with autoregressive PDE surrogates.